\documentclass[10pt]{article} 

\usepackage[accepted]{preamble/rlc}

\usepackage{amssymb}            
\usepackage{mathtools}          
\usepackage{mathrsfs}           
\usepackage{graphicx}           
\usepackage{subcaption}         
\usepackage[space]{grffile}     
\usepackage{url}                

\usepackage{graphicx} 

\usepackage{wrapfig}
\usepackage{xcolor}
\usepackage{graphicx}
\graphicspath{{figures/}}
\usepackage{amsmath}
\usepackage{amsfonts}
\usepackage{cleveref}
\usepackage{svg}
\usepackage{preamble/vector} 
\usepackage{caption}
\usepackage{adjustbox}
\usepackage{longtable,tabularx,booktabs}
\usepackage[flushleft]{threeparttable}
\usepackage{multirow}
\usepackage{lipsum}
\usepackage{makecell}
\usepackage{mdframed}
\usepackage{framed}
\usepackage{comment}
\usepackage{dsfont} 
\usepackage{preamble/vector} 
\usepackage{mathtools}
\usepackage[title]{appendix}
\usepackage[bottom]{footmisc} 
\usepackage{colortbl}
\makeatletter
\@namedef{ver@fixltx2e.sty}{}
\makeatother
\usepackage{dblfloatfix}
\usepackage{hyphenat}
\usepackage{bookmark}
 
\definecolor{juliafunccolor}{HTML}{2e51a2}
\definecolor{juliaparenscolor}{HTML}{f2d71c}

\definecolor{newcolor}{HTML}{000000}
\newcommand{\new}[1]{{\color{newcolor}#1}}

\usepackage{algorithmicx}
\usepackage{algorithm}
\usepackage{algpseudocode}

\algdef{SE}[FOR]{ParallelFor}{EndParallelFor}[1]{\textbf{parallel} \algorithmicfor\ #1}{\algorithmicend}%
\algdef{SE}[FOR]{For}{EndFor}[1]{\algorithmicfor\ #1}{\algorithmicend}%
\algdef{SE}[IF]{If}{EndIf}[1]{\algorithmicif\ #1}{\algorithmicend}%
\algdef{SE}[IF]{NewIf}{NewEndIf}[1]{\new{\algorithmicif}\ #1}{\algorithmicend}%
\algdef{C}[IF]{IF}{ElsIf}[1]{\algorithmicelse\ \algorithmicif\ #1}%
\algdef{Ce}[ELSE]{IF}{Else}{EndIf}{\algorithmicelse}%
\algdef{Ce}[ELSE]{IF}{Else}{NewEndIf}{\algorithmicelse}%
\algdef{SE}[FUNCTION]{Function}{EndFunction}%
   [2]{\algorithmicfunction\ \textproc{#1}\ifthenelse{\equal{#2}{}}{}{(#2)}}%
   {\algorithmicend}%
\algtext*{EndFunction} 
\algrenewcommand\alglinenumber[1]{\color{gray}\tiny #1}

\algtext*{EndFunction} 
\algtext*{EndFor} 
\algtext*{EndParallelFor} 
\algtext*{EndIf} 
\algtext*{NewEndIf} 

\definecolor{commentgray}{rgb}{0.6, 0.6, 0.6}
\newcommand{\GrayComment}[1]{{\hfill{\color{commentgray}$\triangleright$ #1}}}

\usepackage{newfloat}
\usepackage{listings}
\DeclareCaptionStyle{ruled}{labelfont=normalfont,labelsep=colon,strut=off} 
\floatstyle{ruled}
\newfloat{listing}{tb}{lst}{}
\floatname{listing}{Listing}

\title{BetaZero: Belief-State Planning for Long-Horizon POMDPs using Learned Approximations}

\author{%
Robert J. Moss, Anthony Corso, Jef Caers, Mykel J. Kochenderfer\\
Stanford University, \{mossr,\,acorso,\,caers,\,mykel\}@stanford.edu
}

\usepackage{bibentry}

\newcommand{\tworow}[1]{\multirow{2}{*}{{#1}}}

\newcommand{\negphantom}[1]{\settowidth{\dimen0}{#1}\hspace*{-\dimen0}}
\newcommand{\np}{\negphantom{)}}
\newcommand{\lit}[1]{{\color[rgb]{0.7,0.7,0.7}(#1)}\np}
\newcommand{\litdesc}[1]{\lit{#1}\phantom{)}}
\newcommand{\textdoublequotes}{\textsf{''}}

\newcommand{\sameresults}{\textdoublequotes{}}

\usepackage{tikz}
\usepackage{pgfplots}
\pgfplotsset{compat=newest}

\usetikzlibrary{calc}
\usetikzlibrary{shapes.geometric}
\usetikzlibrary{external}
\usetikzlibrary{patterns}
\usetikzlibrary{shapes,arrows,fit}
\usetikzlibrary{positioning}
\usetikzlibrary{arrows.meta, calc, shapes}
\usetikzlibrary{graphs}
\usetikzlibrary{decorations.pathmorphing}
\usetikzlibrary{decorations.pathreplacing}
\usetikzlibrary{backgrounds}
\usetikzlibrary{shadows}

\usepackage{pagecolor}

\definecolor{blues1}{RGB}{198, 219, 239}
\definecolor{blues2}{RGB}{158, 202, 225}
\definecolor{blues3}{RGB}{107, 174, 214}
\definecolor{blues4}{RGB}{49, 130, 189}
\definecolor{blues5}{RGB}{8, 81, 156}

\definecolor{grays1}{RGB}{219, 219, 219}
\definecolor{grays2}{RGB}{202, 202, 202}
\definecolor{grays3}{RGB}{174, 174, 174}
\definecolor{grays4}{RGB}{130, 130, 130}
\definecolor{grays5}{RGB}{81, 81, 81}

\usepackage{pagecolor}

\usetikzlibrary{arrows.meta}
\usetikzlibrary{backgrounds}
\usepgfplotslibrary{patchplots}
\usepgfplotslibrary{fillbetween}
\pgfplotsset{%
    layers/standard/.define layer set={%
        background,axis background,axis grid,axis ticks,axis lines,axis tick labels,pre main,main,axis descriptions,axis foreground%
    }{
        grid style={/pgfplots/on layer=axis grid},%
        tick style={/pgfplots/on layer=axis ticks},%
        axis line style={/pgfplots/on layer=axis lines},%
        label style={/pgfplots/on layer=axis descriptions},%
        legend style={/pgfplots/on layer=axis descriptions},%
        title style={/pgfplots/on layer=axis descriptions},%
        colorbar style={/pgfplots/on layer=axis descriptions},%
        ticklabel style={/pgfplots/on layer=axis tick labels},%
        axis background@ style={/pgfplots/on layer=axis background},%
        3d box foreground style={/pgfplots/on layer=axis foreground},%
    },
}

\pgfplotsset{
colormap={plots1}{rgb(0.00000000)=(0.26700400,0.00487400,0.32941500)
rgb(0.00392157)=(0.26851000,0.00960500,0.33542700)
rgb(0.00784314)=(0.26994400,0.01462500,0.34137900)
rgb(0.01176471)=(0.27130500,0.01994200,0.34726900)
rgb(0.01568627)=(0.27259400,0.02556300,0.35309300)
rgb(0.01960784)=(0.27380900,0.03149700,0.35885300)
rgb(0.02352941)=(0.27495200,0.03775200,0.36454300)
rgb(0.02745098)=(0.27602200,0.04416700,0.37016400)
rgb(0.03137255)=(0.27701800,0.05034400,0.37571500)
rgb(0.03529412)=(0.27794100,0.05632400,0.38119100)
rgb(0.03921569)=(0.27879100,0.06214500,0.38659200)
rgb(0.04313725)=(0.27956600,0.06783600,0.39191700)
rgb(0.04705882)=(0.28026700,0.07341700,0.39716300)
rgb(0.05098039)=(0.28089400,0.07890700,0.40232900)
rgb(0.05490196)=(0.28144600,0.08432000,0.40741400)
rgb(0.05882353)=(0.28192400,0.08966600,0.41241500)
rgb(0.06274510)=(0.28232700,0.09495500,0.41733100)
rgb(0.06666667)=(0.28265600,0.10019600,0.42216000)
rgb(0.07058824)=(0.28291000,0.10539300,0.42690200)
rgb(0.07450980)=(0.28309100,0.11055300,0.43155400)
rgb(0.07843137)=(0.28319700,0.11568000,0.43611500)
rgb(0.08235294)=(0.28322900,0.12077700,0.44058400)
rgb(0.08627451)=(0.28318700,0.12584800,0.44496000)
rgb(0.09019608)=(0.28307200,0.13089500,0.44924100)
rgb(0.09411765)=(0.28288400,0.13592000,0.45342700)
rgb(0.09803922)=(0.28262300,0.14092600,0.45751700)
rgb(0.10196078)=(0.28229000,0.14591200,0.46151000)
rgb(0.10588235)=(0.28188700,0.15088100,0.46540500)
rgb(0.10980392)=(0.28141200,0.15583400,0.46920100)
rgb(0.11372549)=(0.28086800,0.16077100,0.47289900)
rgb(0.11764706)=(0.28025500,0.16569300,0.47649800)
rgb(0.12156863)=(0.27957400,0.17059900,0.47999700)
rgb(0.12549020)=(0.27882600,0.17549000,0.48339700)
rgb(0.12941176)=(0.27801200,0.18036700,0.48669700)
rgb(0.13333333)=(0.27713400,0.18522800,0.48989800)
rgb(0.13725490)=(0.27619400,0.19007400,0.49300100)
rgb(0.14117647)=(0.27519100,0.19490500,0.49600500)
rgb(0.14509804)=(0.27412800,0.19972100,0.49891100)
rgb(0.14901961)=(0.27300600,0.20452000,0.50172100)
rgb(0.15294118)=(0.27182800,0.20930300,0.50443400)
rgb(0.15686275)=(0.27059500,0.21406900,0.50705200)
rgb(0.16078431)=(0.26930800,0.21881800,0.50957700)
rgb(0.16470588)=(0.26796800,0.22354900,0.51200800)
rgb(0.16862745)=(0.26658000,0.22826200,0.51434900)
rgb(0.17254902)=(0.26514500,0.23295600,0.51659900)
rgb(0.17647059)=(0.26366300,0.23763100,0.51876200)
rgb(0.18039216)=(0.26213800,0.24228600,0.52083700)
rgb(0.18431373)=(0.26057100,0.24692200,0.52282800)
rgb(0.18823529)=(0.25896500,0.25153700,0.52473600)
rgb(0.19215686)=(0.25732200,0.25613000,0.52656300)
rgb(0.19607843)=(0.25564500,0.26070300,0.52831200)
rgb(0.20000000)=(0.25393500,0.26525400,0.52998300)
rgb(0.20392157)=(0.25219400,0.26978300,0.53157900)
rgb(0.20784314)=(0.25042500,0.27429000,0.53310300)
rgb(0.21176471)=(0.24862900,0.27877500,0.53455600)
rgb(0.21568627)=(0.24681100,0.28323700,0.53594100)
rgb(0.21960784)=(0.24497200,0.28767500,0.53726000)
rgb(0.22352941)=(0.24311300,0.29209200,0.53851600)
rgb(0.22745098)=(0.24123700,0.29648500,0.53970900)
rgb(0.23137255)=(0.23934600,0.30085500,0.54084400)
rgb(0.23529412)=(0.23744100,0.30520200,0.54192100)
rgb(0.23921569)=(0.23552600,0.30952700,0.54294400)
rgb(0.24313725)=(0.23360300,0.31382800,0.54391400)
rgb(0.24705882)=(0.23167400,0.31810600,0.54483400)
rgb(0.25098039)=(0.22973900,0.32236100,0.54570600)
rgb(0.25490196)=(0.22780200,0.32659400,0.54653200)
rgb(0.25882353)=(0.22586300,0.33080500,0.54731400)
rgb(0.26274510)=(0.22392500,0.33499400,0.54805300)
rgb(0.26666667)=(0.22198900,0.33916100,0.54875200)
rgb(0.27058824)=(0.22005700,0.34330700,0.54941300)
rgb(0.27450980)=(0.21813000,0.34743200,0.55003800)
rgb(0.27843137)=(0.21621000,0.35153500,0.55062700)
rgb(0.28235294)=(0.21429800,0.35561900,0.55118400)
rgb(0.28627451)=(0.21239500,0.35968300,0.55171000)
rgb(0.29019608)=(0.21050300,0.36372700,0.55220600)
rgb(0.29411765)=(0.20862300,0.36775200,0.55267500)
rgb(0.29803922)=(0.20675600,0.37175800,0.55311700)
rgb(0.30196078)=(0.20490300,0.37574600,0.55353300)
rgb(0.30588235)=(0.20306300,0.37971600,0.55392500)
rgb(0.30980392)=(0.20123900,0.38367000,0.55429400)
rgb(0.31372549)=(0.19943000,0.38760700,0.55464200)
rgb(0.31764706)=(0.19763600,0.39152800,0.55496900)
rgb(0.32156863)=(0.19586000,0.39543300,0.55527600)
rgb(0.32549020)=(0.19410000,0.39932300,0.55556500)
rgb(0.32941176)=(0.19235700,0.40319900,0.55583600)
rgb(0.33333333)=(0.19063100,0.40706100,0.55608900)
rgb(0.33725490)=(0.18892300,0.41091000,0.55632600)
rgb(0.34117647)=(0.18723100,0.41474600,0.55654700)
rgb(0.34509804)=(0.18555600,0.41857000,0.55675300)
rgb(0.34901961)=(0.18389800,0.42238300,0.55694400)
rgb(0.35294118)=(0.18225600,0.42618400,0.55712000)
rgb(0.35686275)=(0.18062900,0.42997500,0.55728200)
rgb(0.36078431)=(0.17901900,0.43375600,0.55743000)
rgb(0.36470588)=(0.17742300,0.43752700,0.55756500)
rgb(0.36862745)=(0.17584100,0.44129000,0.55768500)
rgb(0.37254902)=(0.17427400,0.44504400,0.55779200)
rgb(0.37647059)=(0.17271900,0.44879100,0.55788500)
rgb(0.38039216)=(0.17117600,0.45253000,0.55796500)
rgb(0.38431373)=(0.16964600,0.45626200,0.55803000)
rgb(0.38823529)=(0.16812600,0.45998800,0.55808200)
rgb(0.39215686)=(0.16661700,0.46370800,0.55811900)
rgb(0.39607843)=(0.16511700,0.46742300,0.55814100)
rgb(0.40000000)=(0.16362500,0.47113300,0.55814800)
rgb(0.40392157)=(0.16214200,0.47483800,0.55814000)
rgb(0.40784314)=(0.16066500,0.47854000,0.55811500)
rgb(0.41176471)=(0.15919400,0.48223700,0.55807300)
rgb(0.41568627)=(0.15772900,0.48593200,0.55801300)
rgb(0.41960784)=(0.15627000,0.48962400,0.55793600)
rgb(0.42352941)=(0.15481500,0.49331300,0.55784000)
rgb(0.42745098)=(0.15336400,0.49700000,0.55772400)
rgb(0.43137255)=(0.15191800,0.50068500,0.55758700)
rgb(0.43529412)=(0.15047600,0.50436900,0.55743000)
rgb(0.43921569)=(0.14903900,0.50805100,0.55725000)
rgb(0.44313725)=(0.14760700,0.51173300,0.55704900)
rgb(0.44705882)=(0.14618000,0.51541300,0.55682300)
rgb(0.45098039)=(0.14475900,0.51909300,0.55657200)
rgb(0.45490196)=(0.14334300,0.52277300,0.55629500)
rgb(0.45882353)=(0.14193500,0.52645300,0.55599100)
rgb(0.46274510)=(0.14053600,0.53013200,0.55565900)
rgb(0.46666667)=(0.13914700,0.53381200,0.55529800)
rgb(0.47058824)=(0.13777000,0.53749200,0.55490600)
rgb(0.47450980)=(0.13640800,0.54117300,0.55448300)
rgb(0.47843137)=(0.13506600,0.54485300,0.55402900)
rgb(0.48235294)=(0.13374300,0.54853500,0.55354100)
rgb(0.48627451)=(0.13244400,0.55221600,0.55301800)
rgb(0.49019608)=(0.13117200,0.55589900,0.55245900)
rgb(0.49411765)=(0.12993300,0.55958200,0.55186400)
rgb(0.49803922)=(0.12872900,0.56326500,0.55122900)
rgb(0.50196078)=(0.12756800,0.56694900,0.55055600)
rgb(0.50588235)=(0.12645300,0.57063300,0.54984100)
rgb(0.50980392)=(0.12539400,0.57431800,0.54908600)
rgb(0.51372549)=(0.12439500,0.57800200,0.54828700)
rgb(0.51764706)=(0.12346300,0.58168700,0.54744500)
rgb(0.52156863)=(0.12260600,0.58537100,0.54655700)
rgb(0.52549020)=(0.12183100,0.58905500,0.54562300)
rgb(0.52941176)=(0.12114800,0.59273900,0.54464100)
rgb(0.53333333)=(0.12056500,0.59642200,0.54361100)
rgb(0.53725490)=(0.12009200,0.60010400,0.54253000)
rgb(0.54117647)=(0.11973800,0.60378500,0.54140000)
rgb(0.54509804)=(0.11951200,0.60746400,0.54021800)
rgb(0.54901961)=(0.11942300,0.61114100,0.53898200)
rgb(0.55294118)=(0.11948300,0.61481700,0.53769200)
rgb(0.55686275)=(0.11969900,0.61849000,0.53634700)
rgb(0.56078431)=(0.12008100,0.62216100,0.53494600)
rgb(0.56470588)=(0.12063800,0.62582800,0.53348800)
rgb(0.56862745)=(0.12138000,0.62949200,0.53197300)
rgb(0.57254902)=(0.12231200,0.63315300,0.53039800)
rgb(0.57647059)=(0.12344400,0.63680900,0.52876300)
rgb(0.58039216)=(0.12478000,0.64046100,0.52706800)
rgb(0.58431373)=(0.12632600,0.64410700,0.52531100)
rgb(0.58823529)=(0.12808700,0.64774900,0.52349100)
rgb(0.59215686)=(0.13006700,0.65138400,0.52160800)
rgb(0.59607843)=(0.13226800,0.65501400,0.51966100)
rgb(0.60000000)=(0.13469200,0.65863600,0.51764900)
rgb(0.60392157)=(0.13733900,0.66225200,0.51557100)
rgb(0.60784314)=(0.14021000,0.66585900,0.51342700)
rgb(0.61176471)=(0.14330300,0.66945900,0.51121500)
rgb(0.61568627)=(0.14661600,0.67305000,0.50893600)
rgb(0.61960784)=(0.15014800,0.67663100,0.50658900)
rgb(0.62352941)=(0.15389400,0.68020300,0.50417200)
rgb(0.62745098)=(0.15785100,0.68376500,0.50168600)
rgb(0.63137255)=(0.16201600,0.68731600,0.49912900)
rgb(0.63529412)=(0.16638300,0.69085600,0.49650200)
rgb(0.63921569)=(0.17094800,0.69438400,0.49380300)
rgb(0.64313725)=(0.17570700,0.69790000,0.49103300)
rgb(0.64705882)=(0.18065300,0.70140200,0.48818900)
rgb(0.65098039)=(0.18578300,0.70489100,0.48527300)
rgb(0.65490196)=(0.19109000,0.70836600,0.48228400)
rgb(0.65882353)=(0.19657100,0.71182700,0.47922100)
rgb(0.66274510)=(0.20221900,0.71527200,0.47608400)
rgb(0.66666667)=(0.20803000,0.71870100,0.47287300)
rgb(0.67058824)=(0.21400000,0.72211400,0.46958800)
rgb(0.67450980)=(0.22012400,0.72550900,0.46622600)
rgb(0.67843137)=(0.22639700,0.72888800,0.46278900)
rgb(0.68235294)=(0.23281500,0.73224700,0.45927700)
rgb(0.68627451)=(0.23937400,0.73558800,0.45568800)
rgb(0.69019608)=(0.24607000,0.73891000,0.45202400)
rgb(0.69411765)=(0.25289900,0.74221100,0.44828400)
rgb(0.69803922)=(0.25985700,0.74549200,0.44446700)
rgb(0.70196078)=(0.26694100,0.74875100,0.44057300)
rgb(0.70588235)=(0.27414900,0.75198800,0.43660100)
rgb(0.70980392)=(0.28147700,0.75520300,0.43255200)
rgb(0.71372549)=(0.28892100,0.75839400,0.42842600)
rgb(0.71764706)=(0.29647900,0.76156100,0.42422300)
rgb(0.72156863)=(0.30414800,0.76470400,0.41994300)
rgb(0.72549020)=(0.31192500,0.76782200,0.41558600)
rgb(0.72941176)=(0.31980900,0.77091400,0.41115200)
rgb(0.73333333)=(0.32779600,0.77398000,0.40664000)
rgb(0.73725490)=(0.33588500,0.77701800,0.40204900)
rgb(0.74117647)=(0.34407400,0.78002900,0.39738100)
rgb(0.74509804)=(0.35236000,0.78301100,0.39263600)
rgb(0.74901961)=(0.36074100,0.78596400,0.38781400)
rgb(0.75294118)=(0.36921400,0.78888800,0.38291400)
rgb(0.75686275)=(0.37777900,0.79178100,0.37793900)
rgb(0.76078431)=(0.38643300,0.79464400,0.37288600)
rgb(0.76470588)=(0.39517400,0.79747500,0.36775700)
rgb(0.76862745)=(0.40400100,0.80027500,0.36255200)
rgb(0.77254902)=(0.41291300,0.80304100,0.35726900)
rgb(0.77647059)=(0.42190800,0.80577400,0.35191000)
rgb(0.78039216)=(0.43098300,0.80847300,0.34647600)
rgb(0.78431373)=(0.44013700,0.81113800,0.34096700)
rgb(0.78823529)=(0.44936800,0.81376800,0.33538400)
rgb(0.79215686)=(0.45867400,0.81636300,0.32972700)
rgb(0.79607843)=(0.46805300,0.81892100,0.32399800)
rgb(0.80000000)=(0.47750400,0.82144400,0.31819500)
rgb(0.80392157)=(0.48702600,0.82392900,0.31232100)
rgb(0.80784314)=(0.49661500,0.82637600,0.30637700)
rgb(0.81176471)=(0.50627100,0.82878600,0.30036200)
rgb(0.81568627)=(0.51599200,0.83115800,0.29427900)
rgb(0.81960784)=(0.52577600,0.83349100,0.28812700)
rgb(0.82352941)=(0.53562100,0.83578500,0.28190800)
rgb(0.82745098)=(0.54552400,0.83803900,0.27562600)
rgb(0.83137255)=(0.55548400,0.84025400,0.26928100)
rgb(0.83529412)=(0.56549800,0.84243000,0.26287700)
rgb(0.83921569)=(0.57556300,0.84456600,0.25641500)
rgb(0.84313725)=(0.58567800,0.84666100,0.24989700)
rgb(0.84705882)=(0.59583900,0.84871700,0.24332900)
rgb(0.85098039)=(0.60604500,0.85073300,0.23671200)
rgb(0.85490196)=(0.61629300,0.85270900,0.23005200)
rgb(0.85882353)=(0.62657900,0.85464500,0.22335300)
rgb(0.86274510)=(0.63690200,0.85654200,0.21662000)
rgb(0.86666667)=(0.64725700,0.85840000,0.20986100)
rgb(0.87058824)=(0.65764200,0.86021900,0.20308200)
rgb(0.87450980)=(0.66805400,0.86199900,0.19629300)
rgb(0.87843137)=(0.67848900,0.86374200,0.18950300)
rgb(0.88235294)=(0.68894400,0.86544800,0.18272500)
rgb(0.88627451)=(0.69941500,0.86711700,0.17597100)
rgb(0.89019608)=(0.70989800,0.86875100,0.16925700)
rgb(0.89411765)=(0.72039100,0.87035000,0.16260300)
rgb(0.89803922)=(0.73088900,0.87191600,0.15602900)
rgb(0.90196078)=(0.74138800,0.87344900,0.14956100)
rgb(0.90588235)=(0.75188400,0.87495100,0.14322800)
rgb(0.90980392)=(0.76237300,0.87642400,0.13706400)
rgb(0.91372549)=(0.77285200,0.87786800,0.13110900)
rgb(0.91764706)=(0.78331500,0.87928500,0.12540500)
rgb(0.92156863)=(0.79376000,0.88067800,0.12000500)
rgb(0.92549020)=(0.80418200,0.88204600,0.11496500)
rgb(0.92941176)=(0.81457600,0.88339300,0.11034700)
rgb(0.93333333)=(0.82494000,0.88472000,0.10621700)
rgb(0.93725490)=(0.83527000,0.88602900,0.10264600)
rgb(0.94117647)=(0.84556100,0.88732200,0.09970200)
rgb(0.94509804)=(0.85581000,0.88860100,0.09745200)
rgb(0.94901961)=(0.86601300,0.88986800,0.09595300)
rgb(0.95294118)=(0.87616800,0.89112500,0.09525000)
rgb(0.95686275)=(0.88627100,0.89237400,0.09537400)
rgb(0.96078431)=(0.89632000,0.89361600,0.09633500)
rgb(0.96470588)=(0.90631100,0.89485500,0.09812500)
rgb(0.96862745)=(0.91624200,0.89609100,0.10071700)
rgb(0.97254902)=(0.92610600,0.89733000,0.10407100)
rgb(0.97647059)=(0.93590400,0.89857000,0.10813100)
rgb(0.98039216)=(0.94563600,0.89981500,0.11283800)
rgb(0.98431373)=(0.95530000,0.90106500,0.11812800)
rgb(0.98823529)=(0.96489400,0.90232300,0.12394100)
rgb(0.99215686)=(0.97441700,0.90359000,0.13021500)
rgb(0.99607843)=(0.98386800,0.90486700,0.13689700)
rgb(1.00000000)=(0.99324800,0.90615700,0.14393600)},
}

\definecolor{timingcolor}{rgb}{0.55, 0.55, 0.55}
\newcommand{\tcolor}[1]{\color{timingcolor}#1}

\DeclareMathOperator{\Var}{Var}

\DeclareMathOperator*{\argmax}{argmax}

\newcommand{\norm}[1]{\lVert #1 \rVert}
\newcommand*{\defeq}{\stackrel{\text{def}}{=}}
\newcommand{\bpi}{\boldsymbol\pi}

\definecolor{resultcolor}{HTML}{000000}

\newsavebox\CBox
\def\mathBF#1{\sbox\CBox{$#1$}\resizebox{\wd\CBox}{\ht\CBox}{{\color{resultcolor}$\mathbf{#1}$}}}

\usepackage{siunitx}
\makeatletter
\let\amsmath@bigmidtmp\bigm

\newcommand{\bigmid}[1]{%
  \ifcsname fenced@\string#1\endcsname
    \expandafter\@firstoftwo
  \else
    \expandafter\@secondoftwo
  \fi
  {\expandafter\amsmath@bigmidtmp\csname fenced@\string#1\endcsname}%
  {\amsmath@bigmidtmp#1}%
}
\newcommand{\DeclareFence}[2]{\@namedef{fenced@\string#1}{#2}}
\makeatother

\DeclareFence{\mid}{|}

\begin{document}

\maketitle

\begin{abstract}
Real-world planning problems, including autonomous driving and sustainable energy applications like carbon storage and resource exploration, have recently been modeled as partially observable Markov decision processes (POMDPs) and solved using approximate methods. To solve high-dimensional POMDPs in practice, state-of-the-art methods use online planning with problem-specific heuristics to reduce planning horizons and make the problems tractable. Algorithms that learn approximations to replace heuristics have recently found success in large-scale fully observable domains. The key insight is the combination of online Monte Carlo tree search with offline neural network approximations of the optimal policy and value function. In this work, we bring this insight to partially observable domains and propose \textit{BetaZero}, a belief-state planning algorithm for high-dimensional POMDPs. BetaZero learns offline approximations that replace heuristics to enable online decision making in long-horizon problems. We address several challenges inherent in large-scale partially observable domains; namely challenges of transitioning in stochastic environments, prioritizing action branching with a limited search budget, and representing beliefs as input to the network. To formalize the use of all limited search information, we train against a novel $Q$-weighted visit counts policy. We test BetaZero on various well-established POMDP benchmarks found in the literature and a real-world problem of critical mineral exploration. Experiments show that BetaZero outperforms state-of-the-art POMDP solvers on a variety of tasks.\footnote{{Code: \url{https://github.com/sisl/BetaZero.jl}}}
\end{abstract}

\section{Introduction}
Optimizing sequential decisions in real-world settings is challenging due to uncertainties about the true state of the environment.
Modeling such problems as partially observable Markov decision processes (POMDPs) has shown recent success in autonomous driving \citep{wray2021pomdps}, robotics \citep{lauri2022partially}, and aircraft collision avoidance \citep{kochenderfer2012next}.
Solving large or continuous POMDPs require approximations in the form of state-space discretizations or modeling assumptions, e.g., assuming full observability.
Although these approximations are useful when making decisions in a short time horizon, scaling these solutions to long-horizon problems is challenging \citep{shani2013survey}.
Recently, POMDPs have been used to model large-scale information gathering problems such as carbon capture and storage (CCS) \citep{corso2022pomdp,wang2023optimizing}, remediation for groundwater contamination \citep{wang2022sequential}, and critical mineral exploration for battery metals \citep{mern2023intelligent}, and are solved using online tree search methods such as DESPOT \citep{ye2017despot} and POMCPOW \citep{sunberg2018online}.
The performance of these online methods rely on heuristics for action selection (to reduce search tree expansion) and heuristics to estimate the value function (to avoid expensive rollouts and reduce tree search depth).
Without heuristics, online methods have difficulty planning for long-term information acquisition to reason about uncertain future events.
Thus, algorithms to solve high-dimensional POMDPs need to be designed to learn heuristic approximations to enable decision making in long-horizon problems.

\begin{wrapfigure}{r}{0.6\textwidth}
    \centering
    \resizebox{\linewidth}{!}{
        \includegraphics{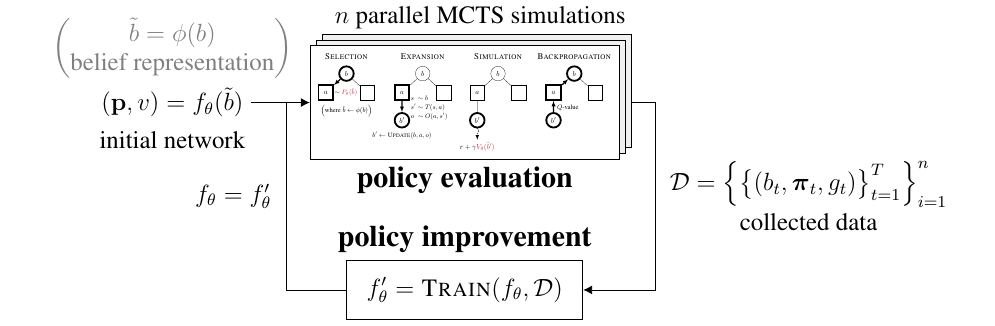}
    }
    \caption{The \textit{BetaZero} POMDP policy iteration algorithm.}
    \label{fig:betazero-alg}
\end{wrapfigure}

\paragraph{Contributions.}
This work aims to address the problem of high-dimensional, long-horizon POMDPs by using the insight of combining online MCTS planning with learned offline neural network approximations that replace heuristics.
Our main contribution is the \textit{BetaZero} belief-state planning algorithm for POMDPs (\cref{fig:betazero-alg}),
addressing the challenges of partial observability in large discrete action spaces and continuous state and observation spaces.
To handle stochastic belief-state transitions, BetaZero uses progressive widening \citep{couetoux2011continuous} to limit belief-state expansion.
When planning in belief space, expensive belief updates \textit{limit the search budget in practice} (e.g., $\mathcal{O}(n)$ for particle filters \citep{thrun2005probabilistic} or $\mathcal{O}(n^3)$ for Kalman filters \citep{welch1995introduction}).
Therefore, we sample from the policy network to prioritize branching on promising actions, and we introduce a novel \textit{$Q$-weighted visit count} policy target that formalizes the use of all information seen during the limited search for policy imitation.
While planning occurs over the full belief, we use a parametric belief representation $\tilde{b} = \phi(b)$ to capture state uncertainty as input to the network.
BetaZero uses the learned policy network $P_\theta(\tilde{b})$ to reduce search breadth and the learned value estimate $V_\theta(\tilde{b})$ to reduce search depth to enable long-horizon online planning (shown in red in \cref{fig:mcts-betazero}).

\section{Problem Formulation}
A partially observable Markov decision process (POMDP) is a model for sequential decision making problems where the true state is unobservable.
Defined by the tuple $\langle \mathcal{S}, \mathcal{A}, \mathcal{O}, T, R, O, \gamma \rangle$, 
POMDPs are an extension to the Markov decision process (MDP) used in reinforcement learning and planning with the addition of an observation space $\mathcal{O}$ (where $o \in \mathcal{O}$) and observation model $O(o \mid a, s')$.
Given a current state $s \in \mathcal{S}$ and taking an action $a \in \mathcal{A}$, the agent transitions to a new state $s'$ using the transition model $s' \sim T(\cdot \mid s, a)$.
Without access to the true state, an observation is received $o \sim O(\cdot \mid a, s')$ and used to update the belief $b$ over the possible next states $s'$ to get the posterior
\begin{equation}    
    b'(s') \propto O(o \mid a, s')\int_{s \mathrlap{\in \mathcal{S}}} T(s' \mid s, a)b(s) \,\mathrm{d}s.
\end{equation}
An example of a type of belief is the non-parametric \textit{particle set} that can represent a broad range of distributions \citep{thrun2005probabilistic}, and \citet{lim2023optimality} show that optimality guarantees exist in finite-sample particle-based POMDP approximations.
Despite choosing to study particle-based beliefs, our work generalizes well to problems with parametric beliefs.

A stochastic POMDP policy $\pi(a \mid b)$ is defined as the distribution over actions given the current belief $b$.
After taking an action $a \sim \pi(\cdot \mid b)$, the agent receives a reward $r$ from the environment according to the reward function $R: \mathcal{S} \times \mathcal{A} \to \mathbb{R}$ or $R: \mathcal{S} \times \mathcal{A} \times \mathcal{S} \to \mathbb{R}$ using the next state.

\paragraph{Belief-state MDPs.}
In \textit{belief-state} MDPs, the POMDP is converted to an MDP by treating the belief as a state \citep{kaelbling1998planning,dmbook}.
The reward function then becomes a weighted sum of the state-based reward:
\begin{align}
    R_b(b,a) = \!\int_{s \mathrlap{\in \mathcal{S}}} b(s)R(s,a) \,\mathrm{d}s \approx \sum_{s \in b} b(s)R(s,a)
\end{align}
The belief-state MDP shares the same action space as the POMDP and operates over a belief space $\mathcal{B}$ that is the simplex over the state space $\mathcal{S}$. The belief-MDP defines a new belief-state transition function $b' \sim T_b(\cdot \mid b, a)$ as:
\begin{align}
    s \sim b(\cdot) \qquad s' \sim T(\cdot \mid s, a) \qquad o \sim O(\cdot \mid a, s') \qquad b' \leftarrow \textsc{Update}(b, a, o) \label{eq:tb_update}
\end{align}
where the belief update can be done using a particle filter \citep{gordon1993novel}.
Therefore, the belief-state MDP is defined by the tuple $\langle \mathcal{B}, \mathcal{A}, T_b, R_b, \gamma \rangle$ with the finite-horizon discount factor $\gamma \in [0,1)$ that controls the effect that future rewards have on the current action.

The objective to solve belief-MDPs is to find a policy $\pi$ that maximizes the \textit{value function}
\begin{equation}
    V^\pi(b_0) = \mathbb{E}_\pi \left[\sum_{t=0}^T \gamma^t R_b(b_t,a_t) \bigmid\mid b_t \sim T_b, a_t \sim \pi \right]
\end{equation}
from an initial belief $b_0$. 
Instead of explicitly constructing a policy over all beliefs,
online planning algorithms estimate the next best action through a planning procedure, often a best-first tree search.

\subsection{Monte Carlo tree search (MCTS)}
Monte Carlo tree search \citep{coulom2007efficient,browne2012survey} is an online, recursive, best-first tree search algorithm to determine the approximately optimal action to take from a given root state of an MDP.
Extensions to MCTS have been applied to POMDPs through several algorithms: \textit{partially observable Monte Carlo planning} (POMCP) treats the state nodes as histories $h$ of action-observation trajectories \citep{silver2010pomcp}, \textit{POMCP with observation widening} (POMCPOW) constructs weighted particle sets at the observation nodes and extends POMCP to fully continuous domains \citep{sunberg2018online}, and \textit{particle filter trees} (PFT) and \textit{information PFT} (IPFT) treat the POMDP as a belief-state MDP and plan directly over the belief-state nodes using a particle filter \citep{ fischer2020information}.
All variants of MCTS execute the following four steps.
In this section we use $s$ to represent the state, the history $h$, and the belief state $b$ and refer to them as ``the state''.

\begin{enumerate}
    \item \textbf{Selection.}\quad 
    During \textit{selection}, an action is selected from the children of a state node based on criteria that balances exploration and exploitation.
    The \textit{upper-confidence tree} algorithm (UCT) \citep{kocsis2006bandit} is a common criterion that selects an action that maximizes the upper-confidence bound $Q(s, a) + c\sqrt{\log N(s)/N(s,a)}$ 
    where $Q(s,a)$ is the $Q$-value estimate for state-action pair $(s,a)$ with a visit count of $N(s,a)$, the total visit count of $N(s) = \sum_a N(s,a)$ for the children ${a \in A(s)}$, and $c$ is an exploration constant.
    \citet{rosin2011multi} introduced the \textit{UCT with predictor} algorithm (PUCT), modified by \citet{silver2017mastering}, where a predictor $P(s,a)$ guides the exploration towards promising branches and selects an action according to the following:
    \begin{equation}
        \argmax_{a \in A(s)}\ Q(s,a) + c\bigg(P(s,a)\frac{\sqrt{N(s)}}{1 + N(s,a)}\bigg) \label{eq:puct}
    \end{equation}

    \item \textbf{Expansion.}\quad 
    In the \textit{expansion} step, the selected action is taken in simulation and the transition model $T(s' \mid s, a)$ is sampled to determine the next state $s'$.
    When the transitions are deterministic, the child node is always a single state.
    If the transition dynamics are stochastic, techniques to balance the branching factor such as progressive widening \citep{couetoux2011continuous} and state abstraction refinement \citep{sokota2021monte} have been proposed.

    \item \textbf{Rollout/Simulation.}\quad 
    In the \textit{rollout} step, also called the \textit{simulation} step due to recursively simulating the MCTS tree expansion, the value is estimated through the execution of a rollout policy until termination or using heuristics to approximate the value function from the given state $s'$.
    Expensive rollouts done by AlphaGo were replaced with a value network lookup in AlphaGo Zero and AlphaZero \citep{silver2016mastering, silver2017mastering, silver2018general}.

    \item \textbf{Backpropagation.}\quad 
    Finally, during the \textit{backpropagation} step, the $Q$-value estimate from the rollout is propagated up the path in the tree as a running average.
\end{enumerate}

\begin{figure}[t!]
    \centering
    \resizebox{0.8\linewidth}{!}{
        \includegraphics{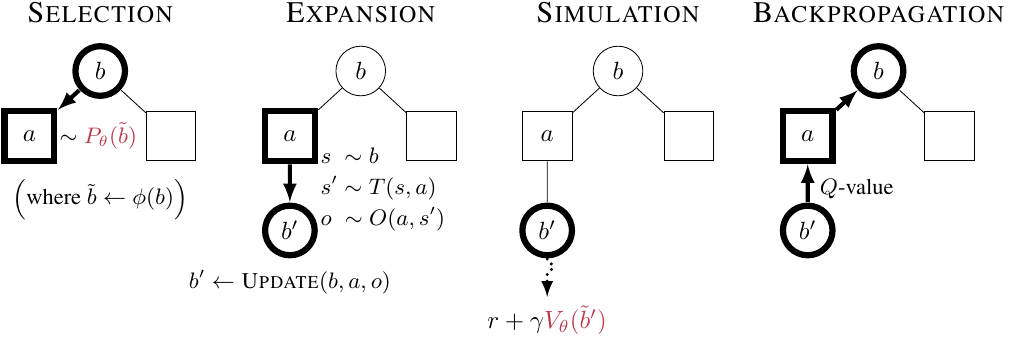}
    }
    \caption{The four stages of MCTS belief-state planning in \textit{BetaZero} using the value $V_\theta$ and policy $P_\theta$ network heads (the \textit{policy evaluation} step in \cref{fig:betazero-alg}).}
    \label{fig:mcts-betazero}
\end{figure}

\paragraph{Root node action selection.}\label{par:root_selection}
After repeating the four steps of MCTS, the final action is selected from the children $a \in A(s)$ of the root state $s$ and executed in the environment.
One way to select the best root node action, referred to as the \textit{robust child} \citep{schadd2009monte, browne2012survey}, selects the action with the highest visit count as $\argmax_a N(s,a)$.
Sampling from the normalized counts, exponentiated by an exploratory temperature $\tau$, is also common \citep{silver2017mastering}.
Another method uses the highest estimated $Q$-value as $\argmax_a Q(s,a)$.
Both criteria have been shown to have problem-based trade-offs \citep{browne2012survey}.

\paragraph{Double progressive widening.}\label{sec:dpw}
To handle stochastic state transitions and large or continuous state and action spaces, double progressive widening (DPW) balances between sampling new nodes to expand on or selecting from existing nodes already in the tree \citep{couetoux2011continuous}.
Two hyperparameters $\alpha \in [0,1]$ and $k \ge 0$ control the branching factor.
If the number of actions tried from state $s$ is less than $kN(s)^\alpha$, then a new action is sampled from the action space and added as a child of node $s$.
Likewise, if the number of expanded states from node $(s,a)$ is less than $kN(s,a)^\alpha$, then a new state is sampled from the transition function $s' \sim T(\cdot \mid s, a)$ and added as a child.
If the state widening condition is not met, then a next state is sampled from the existing children.

Note, in the following sections we will refer to the belief state as $b$ and the true (hidden) state as $s$.

\section{Proposed Algorithm: BetaZero}\label{sec:betazero}

\begin{figure}[t]
    \centering
    \resizebox{0.9\linewidth}{!}{
        \includegraphics{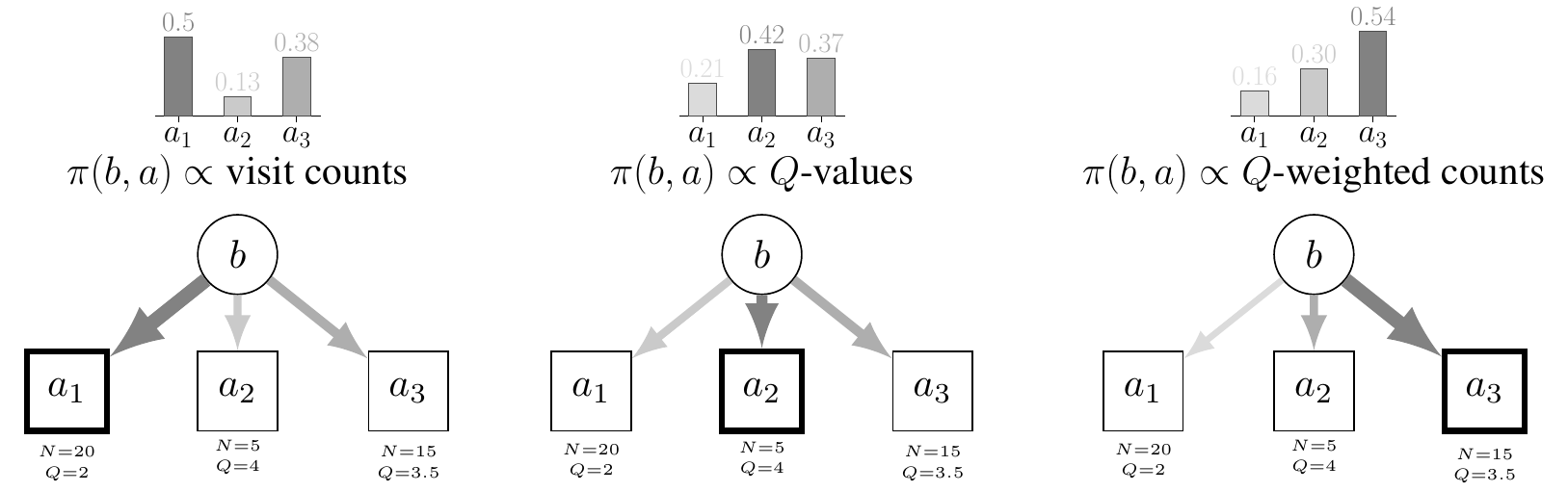}
    }
    \caption{An illustrative example of when collecting policy data based purely on visit counts (left) or $Q$-values (middle) would perform worse than weighting the visit counts based on $Q$-values (right).
    This is useful when using a small MCTS budget with high exploration.
    Using both the $Q$-values \textit{and} visit counts, we incorporate both what the tree search \textit{focused on} and the \textit{values it found}.}
    \label{fig:q-weighting}
\end{figure}

We introduce the \textit{BetaZero} POMDP planning algorithm that replaces heuristics with learned approximations of the optimal policy and value function.
BetaZero is a belief-space policy iteration algorithm with two \textit{offline} steps that learn a network used \textit{online}:
\begin{enumerate}
    \item \textbf{Policy evaluation}: Evaluate the current value and policy network through $n$ parallel episodes of MCTS (\cref{fig:mcts-betazero}) and collect training data: $\mathcal{D} = \left\{\{(b_t, \bpi_t, g_t)\}_{t=1}^T\right\}_{i=1}^n$
    \item \textbf{Policy improvement}: Improve the estimated value function and policy by retraining the neural network parameters $\theta$ with data from the $n_\text{buffer}$ most recent MCTS simulations.
\end{enumerate}
The policy vector over actions $\vect{p} = P_\theta(\tilde{b}, \cdot)$ and the value $v = V_\theta(\tilde{b})$ are combined into a single network with two output heads $(\vect{p}, v) = f_\theta(\tilde{b})$; we refer to $P_\theta$ and $V_\theta$ separately for convenience.
During \textit{policy evaluation}, training data is collected from the outer POMDP loop.
The belief $b_t$ and the tree policy $\bpi_t$ are collected for each time step $t$.
At the end of each episode, the returns $g_t = \sum_{i=t}^T \gamma^{(i-t)} r_i$ are computed from the set of observed rewards for all time steps up to a terminal horizon $T$.
Traditionally, MCTS algorithms use a tree policy $\bpi_t$ that is proportional to the root node visit counts of its children actions $\bpi_t(b_t, a) \propto N(b_t,a)^{1/\tau}$.
The counts are sampled after exponentiating with a temperature $\tau$ to encourage exploration but evaluated online with $\tau \to 0$ to return the maximizing action \citep{silver2017mastering}.
In certain settings, relying solely on visit counts may overlook crucial information (see \cref{fig:q-weighting}).

\paragraph{Policy vector as \texorpdfstring{$Q$}{Q}-weighted counts.}
When planning in belief space, expensive belief updates occur in the tree search and thus may limit the MCTS budget.
Therefore, the visit counts may not converge towards an optimal strategy as the budget may be spent on exploration.
\citet{danihelka2022policy} and \citet{czech2021improving} suggest using knowledge of the $Q$-values from search in MCTS action selection.
Using only tree information, we incorporate $Q$-values and train against the policy 
\begin{equation}
    \bpi_t(b_t, a) \propto \Biggl(\biggl(\frac{\exp Q(b_t, a)}{\sum_{a'} \exp Q(b_t, a')}\biggr)^{z_q}\biggl(\frac{N(b_t,a)}{\sum_{a'} N(b_t,a')}\biggr)^{z_n}\Biggr)^{1/\tau}\label{eq:policy_q_weight}
\end{equation}
which is then normalized to get a valid probability distribution.
\Cref{eq:policy_q_weight} simply weights the visit counts by the softmax $Q$-value distribution with parameters $z_q \in [0,1]$ and $z_n \in [0,1]$ defining the influence of the values and the visit counts, respectively.
If $z_q=z_n=1$, then the influence is equal and if $z_q=z_n=0$, then the policy becomes uniform.
Once the tree search finishes, the root node action is selected from $a \sim \bpi_t(b_t, \cdot)$ and returns the argmax when the temperature $\tau \to 0$.

\paragraph{Loss function.}
Using the latest collected data, the \textit{policy improvement} step retrains the policy network head using the cross-entropy loss $\mathcal{L}_{P_\theta}(\bpi_t, \vect{p}_t) = -\bpi_t^\top \log \vect{p}_t$.
The value network head is simultaneously trained to fit the returns $g_t$ using mean-squared error (MSE) or mean-absolute error (MAE) to predict the value of the belief $b_t$.
Note that we use either MSE or MAE value losses $\mathcal{L}_{V_\theta}$ for different problems depending on the characteristics of the return distribution.
In sparse reward problems, MAE is a better choice as the distribution is closer to Laplacian \citep{hodson2022root}.
When the reward is distributed closer to Gaussian, then MSE is more suitable \citep{chai2014root}.
The final loss function combines the value and policy losses with $L_2$-regularization scaled by $\lambda$:
\begin{equation}
    \ell_{\beta_0} = \mathcal{L}_{V_\theta}(g_t, v_t) + \mathcal{L}_{P_\theta}(\bpi_t, \vect{p}_t) + \lambda\norm{\theta}^2
\end{equation}

\paragraph{Prioritized action widening.}
Planning in belief space explicitly handles state uncertainty but may incur computational overhead when performing belief updates, therefore we avoid trying all actions at every belief node.
We apply \textit{action progressive widening} \citep{couetoux2011continuous} to limit action expansion, which has been used in the context of continuous action spaces \citep{moerland2018a0c} and large discrete action spaces \citep{yee2016monte}. 
\citet{browne2012survey} found action progressive widening to be effective in cases where favorable actions were tried first and \citet{mern2021improved} show that prioritizing actions can improve MCTS performance in large discrete action spaces. 
Therefore, BetaZero selects actions through progressive widening and uses information from the learned policy network to sample new actions $a \sim P_\theta(\tilde{b}, \cdot)$, line \ref{line:sample_action}, alg. \ref{alg:betazero-action-pw}.
This way, we first focus the expansion on \textit{promising actions}, then make the final selection based on PUCT.\footnote{PUCT uses normalized $Q$-values from $0$ to $1$ ($\bar{Q}$) so $c$ can be problem independent \citep{schrittwieser2020mastering}.}
In \cref{sec:experiments}, we perform an ablation to measure the effect of using the policy $P_\theta$ to prioritize actions when widening the tree.

\resizebox{0.49\textwidth}{!}{%
    \begin{minipage}{0.58\textwidth}
        \begin{algorithm}[H]
    \small
    \caption{BetaZero action progressive widening.}
    \label{alg:betazero-action-pw}
    \begin{algorithmic}[1]
        \Function{ActionSelection$(\new{f_\theta}, b)$}{}
            \State \new{$\tilde{b} \leftarrow \phi(b)$} \GrayComment{belief representation}
            \NewIf {\new{$| A(b) | \le k_a N(b)^{\alpha_a}\phantom{\tilde{b}}$}} \GrayComment{action progressive widening}
                \State \new{$a \sim P_\theta(\tilde{b}, \cdot)$} \GrayComment{prioritized from network}\label{line:sample_action}
                \State $N(b,a) \leftarrow N_0(b,a)$
                \State $Q(b,a) \leftarrow Q_0(b,a)$ \label{line:init_q} \GrayComment{bootstrap initial $Q$-value}
                \State $A(b) \leftarrow A(b) \cup \{a\}$ \GrayComment{add to visited actions $A(b)$}
            \NewEndIf
            \State \Return \new{${\displaystyle\argmax_{a \in A(b)}}\ \bar{Q}(b,a) + c\Bigl(P_\theta(\tilde{b},a)\frac{\sqrt{N(b)}}{1 + N(b,a)}\Bigr)$}
        \EndFunction
    \end{algorithmic}
\end{algorithm}

    \end{minipage}
}
\resizebox{0.49\textwidth}{!}{%
    \begin{minipage}{0.58\textwidth}
        \begin{algorithm}[H]
    \small
    \caption{BetaZero belief-state progressive widening.}
    \label{alg:betazero-state-pw}
    \begin{algorithmic}[1]
        \Function{BeliefStateExpansion$(b,a)$}{}
            \NewIf {\new{$|B(b,a)| \le k_b N(b,a)^{\alpha_b}$}} \GrayComment{belief progressive widening}
                \State $b' \sim T_b(\cdot \mid b, a)$ \GrayComment{\cref{eq:tb_update}}
                \State $B(b,a) \leftarrow B(b,a) \cup \{b'\}$ \GrayComment{add to visited beliefs}
            \Else
                \State $b' \sim B(b,a)$ \GrayComment{sample from belief-states in the tree}
            \NewEndIf
            \State $r \leftarrow R(b, a)$ or $r \leftarrow R(b, a, b')$
            \State \Return $b', r$
        \EndFunction
    \end{algorithmic}
\end{algorithm}
    \end{minipage}
}

\paragraph{Stochastic belief-state transitions.}
A challenge with~partially observable domains is handling non-deterministic belief-state transitions in the tree search.
The belief-state transition function $T_b$ consists of several stochastic components and the belief is continuous (being a probability distribution over states).
To address this, we use progressive widening from \citet{couetoux2011continuous} (\cref{alg:betazero-state-pw}).
Other methods for state expansion, like \textit{state abstraction refinement} from \citet{sokota2021monte}, rely on problem-specific distance metrics between states to perform a nearest neighbor search.
Progressive widening avoids problem-specific heuristics by using information only available in the search tree to provide artificially limited belief-state branching.
Limited branching is important as the belief updates can be computationally expensive, thus limiting the MCTS search budget in practice.

\paragraph{Parametric belief representation.}
Inputting state histories into the network has been done in the literature, in both the context of MDPs \citep{silver2018general} and POMDPs \citep{cai2022closing}
Using only state information does not generalize to complex POMDPs (seen in \cref{fig:ablations}), therefore, a representation of the belief is required.
Although a particle belief is not parametrically defined, approximating the belief as summary statistics (e.g., mean and std) may capture enough information for value and policy estimation to be used during planning \citep{coquelin2008particle}.
Approximating the particle set parametrically is easy to implement and computationally inexpensive.
We show that the approximation works well across various problems and, unsurprisingly, using only the mean state is inadequate (see \cref{sec:experiments}).
We represent the particle set $b$ parametrically as $\phi(b) = [\mu(b), \sigma(b)]$. 
BetaZero plans over the full belief $b$ in the tree and only converts to the belief representation $\tilde{b} = \phi(b)$ for network evaluations.
We do not depend on the \textit{exact} way in which the belief is represented, so long as it captures state uncertainty.
\citet{coquelin2008particle} consider how to represent a particle filter belief as a finite set of features for policy gradient and suggest the approximation that consists of the mean and covariance, but only consider the class of policies depending on a single feature of the mean.
Their work suggests that other features, such as entropy, could also be used.
Other algorithms (e.g., FORBES from \citet{chen2022flow}) could instead be used to learn this belief representation.
Another example approach could use principle component analysis (PCA) to learn lower-dimensional features for belief representation \citep{roy2005finding}.

\paragraph{Bootstrapping initial \texorpdfstring{$Q$}{Q}-values.} The value network $V_\theta$ is used during the \textit{simulation} step to replace rollouts with a network lookup (line \ref{line:mcts_lookup}, alg. \ref{alg:betazero-mcts}).
When a new state-action node is added to the tree, initial $Q$-values can also use the value network to bootstrap the estimate:
\begin{equation}
    Q_0(b, a) \defeq R_b(b, a) + \gamma V_\theta(\phi(b')) \ \  \text{where} \ \  b' \sim T_b(\cdot \mid b, a)
\end{equation}
Bootstrapping occurs in \cref{alg:betazero-action-pw} (line \ref{line:init_q}) and incurs an additional belief update through the belief-state transition $T_b$ and may be opted only during online execution.
The bootstrapped estimate is more robust \citep{kumar2019stabilizing} and can be useful to initialize online search.
Note that bootstrapping is also used in the model-free \textit{MuZero} algorithm \citep{schrittwieser2020mastering}.

\paragraph{Complexity analysis.}
The runtime complexity of MCTS is $M = \mathcal{O}(ndm)$ for the $n$ number of MCTS iterations (denoted $n_\text{online}$ in \cref{alg:betazero-mcts}), for the search depth $d$, and with a belief update over $m$ particles at each belief-state node.
The full complexity of BetaZero is $\mathcal{O}(pmTM/c)$ for $p$ parallel runs (denoted $n_\text{data}$ in \cref{alg:collect_data}), an episode horizon of $T$ (each step updating the belief over $m$ particles), the MCTS complexity of $M$, and the number of CPU cores $c$.
\begin{wrapfigure}{r}{0.4\textwidth}
    \resizebox{0.4\textwidth}{!}{%
        \begin{minipage}{0.44\textwidth}
            \input{algorithms/mcts}
        \end{minipage}
    }
    \vspace{-3mm}
\end{wrapfigure}%
The memory complexity for MCTS is $E = \mathcal{O}(k^d)$ for $k = |A(b)||B(b,a)|$ where 
$|B(b,a)|$ is the number of belief-action nodes and $|A(b)|$ is the number of children, which depend on progressive widening parameters.
The memory complexity for BetaZero is $\mathcal{O}(TPE|\theta|)$ for the collected data sizes of the belief and returns $T$ (same as the horizon), the policy vector size of $P = |\mathcal{A}|$ (i.e., action space size), the MCTS memory complexity of $E$, and the network size of $|\theta|$.
Compared to standard MCTS applications to belief-state MDPs, BetaZero requires additional memory for data collection and neural network storage.

\Cref{alg:betazero-mcts} details MCTS for BetaZero with extensions for belief-state planning with learned approximations. The full BetaZero algorithm is shown in \cref{alg:betazero,alg:collect_data,alg:mcts-top-lvl}.

\section{Related Work}
Algorithms to solve high-dimensional, \textit{fully observable} Markov decision processes (MDPs) have been proposed to learn approximations that replace problem-specific heuristics.
\citet{silver2018general} introduced the \textit{AlphaZero} algorithm for large, deterministic MDPs and showed considerable success in games such as Go, chess, shogi, and Atari \citep{silver2018general, schrittwieser2020mastering}.
The success is attributed to the combination of online Monte Carlo tree search (MCTS) and a neural network that approximates the optimal value function and the offline policy.
Extensions of AlphaZero and the model-free variant \textit{MuZero} \citep{schrittwieser2020mastering} have already addressed several challenges when applying to broad classes of MDPs.
For large or continuous action spaces, \citet{hubert2021learning} introduced a policy improvement algorithm called \textit{Sampled MuZero} that samples an action set of an \textit{a priori} fixed size every time a node is expanded.
\citet{antonoglou2021planning} introduced \textit{Stochastic MuZero} that extends MuZero to games with stochastic transitions but assumes a finite set of possible next states so that each transition can be associated with a chance outcome.
Applying these algorithms to large or continuous spaces with partially observability remains challenging.

To handle partial observability in stochastic games, \citet{ozair2021vector} combine VQ-VAEs with MuZero to encode future discrete observations into latent variables.
Other approaches handle partial observability by inputting action-observation histories directly into the network \citep{kimura2020development, vinyals2019grandmaster}.
Similarly, \citet{igl2018deep} introduce a method to learn a belief representation within the network when the agent is only given access to histories.
Their work focuses on the \textit{reinforcement learning} (RL) domain and they show that a belief distribution can be represented as a latent state in the learned model.
The FORBES algorithm \citep{chen2022flow} builds a normalizing flow-based belief and learns a policy through an actor-critic RL algorithm.
Methods to learn the belief are necessary when a prior belief model is not available.
When such models \textit{do} exist, as is the case with many POMDPs that we study, using the models can be valuable for long-term planning.
\citet{hoel2019combining} apply AlphaGo Zero \citep{silver2017mastering} to an autonomous driving POMDP using the most-likely state as the network input but overlook significant belief uncertainty information.

\paragraph{Planning vs. reinforcement learning.}
In POMDP \textit{planning}, models of the transitions, rewards, and observations are known.
In contrast, in the model-based partially observable reinforcement learning (PORL) domain, these models are learned along with a policy or value function \citep{sutton2018reinforcement,subramanian2022approximate}.
A difference between these settings is that PORL algorithms reset the agent and learn through experience, while planning algorithms, like MCTS, must consider future trajectories from any state.
When RL problems have deterministic state transitions, they can be cast as a planning problem by replaying the full state trajectory along a tree path, which may be prohibitively expensive for long-horizon problems.
Both settings are closely related and pose interesting research challenges.
Specifically, sequential planning over given models in high-dimensional, long-horizon POMDPs remains challenging \citep{lauri2022partially}.

\paragraph{Online POMDP planning.}
\citet{sunberg2018online} introduced the \textit{POMCPOW} planning algorithm that iteratively builds a particle set belief within the tree, designed for fully continuous spaces.
In practice, POMCPOW relies on heuristics for value function estimation and action selection (e.g., work from \citet{mern2023intelligent}).
\citet{wu2021adaptive} introduced \textit{AdaOPS} that adaptively approximates the belief through particle filtering and maintains value function bounds that are initialized with heuristics (e.g., solving the MDP or using expert policies).
The major limitation of existing solvers is the reliance on heuristics to make long-horizon POMDPs tractable, which may not scale to high-dimensional problems.
\citet{cai2022closing} proposed \textit{LeTS-Drive} applied to autonomous driving that combines planning and learning similar to BetaZero, and uses HyP-DESPOT with PUCT exploration \citep{cai2021hyp} as the planning algorithm, instead of MCTS.
It uses a state-history window as input to the network, which may not adequately capture the state uncertainty.
LeTS-Drive expands on all actions during planning, which we show may lead to suboptimal planning under limited search budgets (shown in \cref{fig:apw_sensitivity_ld10,fig:apw_sensitivity_rs20}).
To handle long-horizon POMDPs, \citet{mazzi2023learning} propose learning logic-based rules as policy guidance in POMCP, yet domain-specific knowledge is required to define the set of features for the rules, which may not be easily generalized to complex POMDPs we study in this work.
Therefore, we identified the need for a general POMDP planning algorithm that does not rely on problem-specific heuristics for good performance.

\section{Experiments}\label{sec:experiments}
\begin{wrapfigure}{r}{0.4\textwidth}
    \centering
    \begin{threeparttable}
      \begin{adjustbox}{max width=0.4\textwidth}
      \begin{small}
        \begin{tabular}{@{}lrrr@{}}
            \toprule
                                                   &  $|\mathcal{S}|$              &  $|\mathcal{A}|$  &  $|\mathcal{O}|$  \\
            \midrule
            $\text{LightDark}(5 \text{ and } 10)$  &  $|\mathbb{R}|$               &  $3$              &  $|\mathbb{R}|$  \\
            $\text{RockSample}(15,15)$             &  $7{,}372{,}800$              &  $20$             &  $3$  \\
            $\text{RockSample}(20,20)$             &  $419{,}430{,}400$            &  $25$             &  $3$  \\
            Mineral Exploration                    &  $|\mathbb{R}^{32\times32}|$  &  $38$             &  $|\mathbb{R}_{\ge 0}|$  \\
            \bottomrule
        \end{tabular}
        \end{small}
    \end{adjustbox}
    \end{threeparttable}
    \caption{POMDP space dimensions.}\label{tab:spaces}
\end{wrapfigure}
Three benchmark problems were chosen to evaluate the performance of BetaZero.
\Cref{tab:spaces} details the POMDP sizes and appendices further describe the POMDPs, network architectures, and experimental design.

In \textsc{LightDark$(y)$} from \citet{platt2010belief}, the goal of the agent is to execute a \texttt{stop} action at the origin while receiving noisy observations of its true location.
The noise is minimized in the ``light'' region ${y=5}$.
We also benchmark against a more challenging version with the light region at ${y=10}$ from \citet{sunberg2018online}, and restrict the agent to only three actions: move \texttt{up} or \texttt{down} by one, or \texttt{stop}.
The modified problem requires information gathering over longer horizons.
Next is the \textsc{RockSample$(n,k)$} POMDP \citep{smith2012heuristic}, which is a scalable information gathering problem where an agent moves in an $n \times n$ grid to observe $k$ rocks with an objective to sample only the ``good'' rocks.
Well-established POMDP benchmarks go up to $n=15$ and $k=15$; we also test a harder version with $n=20$ and $k=20$ to show the scalability of BetaZero, noting that this case has been evaluated in the multi-agent setting \citep{cai2021hyp}.
Finally, in the real-world \textsc{Mineral Exploration} problem \citep{mern2023intelligent}, the agent drills over a $32\times32$ region to determine if a subsurface ore body should be mined or abandoned and the continuous ore quality is observed at the drill locations to build a belief.
Drilling incurs a penalty, and if chosen to mine, then the agent is rewarded or penalized  based on an economic threshold of the extracted ore mass.
The problem is challenging due to reasoning over limited observations with sparse rewards.

\begin{table*}[t!]
    \centering
    \begin{threeparttable}
        \renewcommand{\arraystretch}{0.95}
        \begin{adjustbox}{max width=\textwidth}
        \begin{tabular}{@{}lrrrrrrrrrr@{}}
            \arrayrulecolor{black} 
            \toprule
                &  \multicolumn{2}{c}{$\text{LightDark}(5)$}  &  \multicolumn{2}{c}{$\text{LightDark}(10)$}  &  \multicolumn{2}{c}{$\text{RockSample}(15,15)$}  &  \multicolumn{2}{c}{$\text{RockSample}({20,20})$}  &  \multicolumn{2}{c}{Mineral Exploration} \\
            \arrayrulecolor{black}
            \cmidrule{2-11}
            \arrayrulecolor{black} 
                &  returns  &  \tcolor time [\textit{off},\textit{on}] s  &  returns  &  \tcolor time [\textit{off},\textit{on}] s  &  returns  &  \tcolor time [\textit{off},\textit{on}] s  &  returns  &  \tcolor time [\textit{off},\textit{on}] s  &  returns  &  \tcolor time [\textit{off},\textit{on}] s \\
            \midrule
            \arrayrulecolor{white}
            BetaZero  &  $\mathBF{4.47 \pm 0.28}$  &  \tcolor{$[\num{2274},\,\num{0.014}]$}  &  $\mathBF{16.77 \pm 1.28}$  &  \tcolor{$[\num{2740},\,\num{0.331}]$}  &  $\num{20.15 \pm 0.71}$  &  \tcolor{$[\num{5701},\,\num{0.477}]$}  &  $\mathBF{13.09 \pm 0.55}$  &  \tcolor{$[\num{7081},\,\num{1.109}]$}  &  $\mathBF{10.67 \pm 2.25}$  &  \tcolor{$[\num{22505},\,\num{5.126}]$}  \\
            \midrule
            Raw Policy $P_\theta$  &  $\num{4.44 \pm 0.28}$  &  \tcolor{$[\num{2274},\,\num{0.004}]$}  &  $\num{13.74 \pm 1.33}$  &  \tcolor{$[\num{2740},\,\num{0.004}]$}  &  $\num{10.96 \pm 0.98}$  &  \tcolor{$[\num{5701},\,\num{0.018}]$}  &  $\num{2.03 \pm 0.34}$  &  \tcolor{$[\num{7081},\,\num{0.084}]$}  &  $\num{8.67 \pm 2.52}$  &  \tcolor{$[\num{22505},\,\num{0.533}]$}  \\
            \midrule
            Raw Value $V_\theta$\tnote{*}  &  $\num{3.16 \pm 0.4}$  &  \tcolor{$[\num{2274},\,\num{0.008}]$}  &  $\num{12.7 \pm 1.46}$  &  \tcolor{$[\num{2740},\,\num{0.009}]$}  &  $\num{9.96 \pm 0.65}$  &  \tcolor{$[\num{5701},\,\num{0.158}]$}  &  $\num{3.57 \pm 0.40}$  &  \tcolor{$[\num{7081},\,\num{0.204}]$}  &  $\num{9.75 \pm 2.42}$  &  \tcolor{$[\num{22505},\,\num{1.420}]$}  \\
            \arrayrulecolor{black}\midrule
            \tworow{AdaOPS}  &  $\num{3.78 \pm 0.27}$  &  \tworow{\tcolor{$[\num{68},\,\num{0.089}]$}}  &  \tworow{$\num{5.22 \pm 1.77}$}  &  \tworow{\tcolor{$[\num{81},\,\num{0.510}]$}}  &  $\mathBF{20.67 \pm 0.72}$  &  \tworow{\tcolor{$[\num{7},\,\num{2.768}]$}}  &  \tworow{---}  &  \tworow{---}  &  \tworow{$\num{3.33 \pm 1.95}$}  &  \tworow{\tcolor{$[\num{5},\,\num{0.112}]$}}  \\
                             &  \lit{$\num{3.79 \pm 0.07}$}  &  &  &  & \lit{$\num{17.16 \pm 0.21}$}  &  &  &  &  &  \\
            \arrayrulecolor{white}\midrule
            AdaOPS (fixed bounds)  &  $\num{3.7 \pm 0.25}$  &  \tcolor{$[\num{0},\,\num{0.039}]$}  &  $\num{4.98 \pm 2.01}$  &  \tcolor{$[\num{0},\,\num{0.573}]$}  &  $\num{13.37 \pm 0.71}$  &  \tcolor{$[\num{0},\,\num{1.349}]$}  &  $\num{11.66 \pm 0.49}$  &  \tcolor{$[\num{1},\,\num{1.458}]$}  &  \sameresults  &  \sameresults  \\
            \arrayrulecolor{grays1}\midrule
            \tworow{POMCPOW}  &  $\num{3.21 \pm 0.38}$  &  \tworow{\tcolor{$[\num{59},\,\num{0.189}]$}}  &  \tworow{$\num{0.68 \pm 0.41}$}  &  \tworow{\tcolor{$[\num{70},\,\num{1.261}]$}}  &  $\num{11.14 \pm 0.59}$  &  \tworow{\tcolor{$[\num{0},\,\num{0.929}]$}}  &  \tworow{$\num{10.22 \pm 0.47}$}  &  \tworow{\tcolor{$[\num{0},\,\num{1.480}]$}}  &  \tworow{$\num{9.43 \pm 2.19}$}  &  \tworow{\tcolor{$[\num{0},\,\num{6.728}]$}}  \\
                              &  \lit{$\num{3.23 \pm 0.11}$}  &  &  &  &  \lit{$\num{10.40 \pm 0.18}$}  &  &  &  &  &  \\
            \arrayrulecolor{white}\midrule
            POMCPOW (no heuristics)  &  $\num{1.96 \pm 0.58}$  &  \tcolor{$[\num{0},\,\num{0.099}]$}  &  $\num{-5.9 \pm 5.78}$  &  \tcolor{$[\num{0},\,\num{0.742}]$}  &  $\num{10.17 \pm 0.61}$  &  \tcolor{$[\num{0},\,\num{1.485}]$}  &  $\num{4.03 \pm 0.44}$  &  \tcolor{$[\num{0},\,\num{5.173}]$}  &  $\num{5.38 \pm 2.15}$  &  \tcolor{$[\num{0},\,\num{5.915}]$}  \\
            \arrayrulecolor{grays1}\midrule
            \tworow{DESPOT}  &  $\num{2.37 \pm 0.37}$  &  \tworow{\tcolor{$[\num{0},\,\num{0.008}]$}}  &  \tworow{$\num{0.43 \pm 0.36}$}  &  \tworow{\tcolor{$[\num{0},\,\num{0.046}]$}}  &  $\num{18.44 \pm 0.69}$  &  \tworow{\tcolor{$[\num{7},\,\num{3.822}]$}}  &  \tworow{---}  &  \tworow{---}  &  \tworow{$\num{5.29 \pm 2.17}$}  &  \tworow{\tcolor{$[\num{5},\,\num{0.283}]$}}  \\
                             &  \lit{$\num{2.50 \pm 0.10}$}  &  &  &  &  \lit{$\num{15.67 \pm 0.20}$}  &  &  &  &  &  \\
            \arrayrulecolor{white}\midrule
            DESPOT (fixed bounds)  &  $\num{2.70 \pm 0.50}$  &  \tcolor{$[\num{0},\,\num{0.008}]$}  &  $\num{0.49 \pm 0.30}$  &  \tcolor{$[\num{0},\,\num{0.025}]$}  &  $\num{4.29 \pm 0.45}$  &  \tcolor{$[\num{0},\,\num{5.091}]$}  &  $\num{0.00 \pm 0.00}$  &  \tcolor{$[\num{0},\,\num{5.179}]$}  &  \sameresults  &  \sameresults \\
            \arrayrulecolor{grays1}\midrule
            \arrayrulecolor{white}\midrule
            LeTS-Drive (HyP-DESPOT + $f_\theta$)  &  $\num{3.05 \pm 0.25}$  &  \tcolor{$[\num{1260},\,\num{0.019}]$}  &  $\num{4.08 \pm 5.48}$  &  \tcolor{$[\num{1529},\,\num{0.058}]$}  &  $\num{11.22 \pm 0.27}$  &  \tcolor{$[\num{48064},\,\num{1.576}]$}  &  $\num{9.68 \pm 0.25}$  &  \tcolor{$[\num{63850},\,\num{2.018}]$}  &  $\num{3.17 \pm 2.04}$  &  \tcolor{$[\num{11738},\,\num{4.613}]$}  \\
            \arrayrulecolor{white}\midrule
            \arrayrulecolor{black}\midrule
            Approx. Optimal  &  $\num{4.06 \pm 0.31}$  &  \tcolor{$[\num{18359},\,\num{0.094}]$}  &  $\num{15.04 \pm 1.27}$  &  \tcolor{$[\num{19548},\,\num{0.024}]$}  &  ---  &  ---  &  ---  &  ---  &  $\num{11.9 \pm 0.18}$  &  N/A  \\
            \arrayrulecolor{black} 
            \bottomrule
        \end{tabular}
        \end{adjustbox}
        \begin{scriptsize}
            \begin{tablenotes}
                \item[*] {One-step look-ahead over all actions using only the value network with $5$ observations per action.}
                \item[\phantom{$\dagger$}] {Entries with ``---'' failed to run, \textdoublequotes{} are the same as the ones above, and entries in \litdesc{parentheses} are from the literature.}
            \end{tablenotes}
        \end{scriptsize}
    \end{threeparttable}
    \caption{Results comparing \textit{BetaZero} to various state-of-the-art POMDP solvers. Reporting return mean and standard error over \num{100} seeds, and [\textit{offline}, \textit{online}] timing in seconds.}\label{tab:results}
\end{table*}

\begin{figure}[b!]
    \centering
    \includegraphics[width=0.8\linewidth]{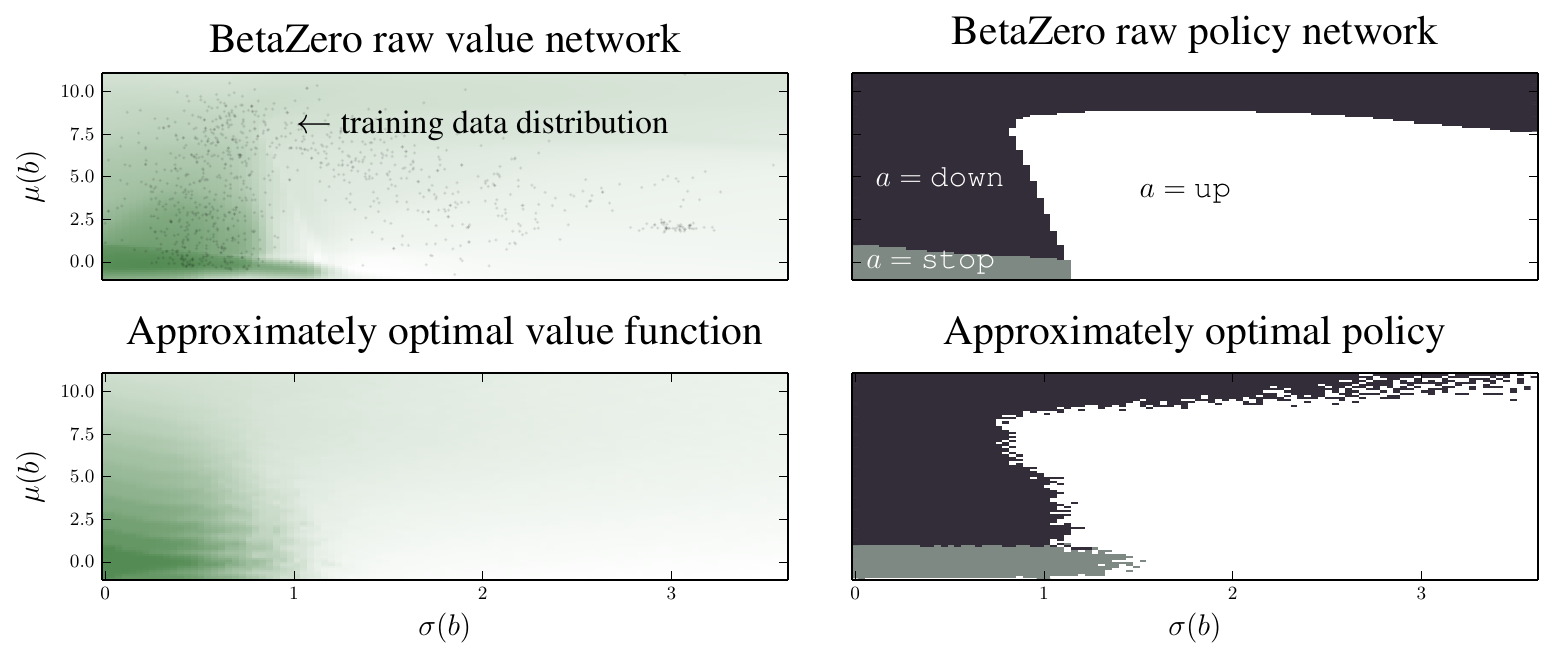}
    \caption{\textsc{LightDark}$(10)$ value and policy plots over belief mean and std. High uncertainty (horizontal axis) makes the agent localize \texttt{up} near $y=10$, then moves \texttt{down} and \texttt{stops} at the origin.}
    \label{fig:lightdark_value_policy}
\end{figure}

We baseline BetaZero against several online POMDP algorithms, namely AdaOPS, POMCPOW, DESPOT, and LeTS-Drive (HyP-DESPOT with a learned network).
In LightDark, we solve for an approximately optimal policy using \textit{local approximation value iteration} (LAVI) \citep{dmubook} over a discretized parametric belief space, and for mineral exploration, the value estimates come from privileged information described in the appendix.
For a fair comparison, parameters were set to roughly match the total number of simulations of about one million per algorithm.

\subsection{Empirical results and discussion}
\Cref{tab:results} shows that BetaZero outperforms state-of-the-art algorithms in most cases, with larger improvements when baseline algorithms do not rely on heuristics.
While BetaZero has a large offline timing component, similar to LeTS-Drive, it is significantly less than solving for the approximately optimal policy.
\Cref{fig:lightdark_value_policy} compares the raw BetaZero value and policy network with \textit{value iteration} for \textsc{LightDark}$(10)$.
Qualitatively, BetaZero learns an accurate optimal policy and value function in areas where training data was collected.
Areas where BetaZero and the approximately optimal policy diverge may be a result of a lack of training data in those regions (top right corners).
Despite this, BetaZero remains nearly optimal as those beliefs do not occur during execution.
Out-of-distribution methods could quantify this uncertainty, e.g., an ensemble of networks \citep{salehi2022unified}.

In \textsc{RockSample}$(15,15)$, BetaZero is comparable to AdaOPS yet scales better to higher dimensional problems such as the \textsc{RockSample}$(20,20)$ POMDP.
AdaOPS computes an upper bound using QMDP \citep{littman1995learning} to find the optimal utility of the fully observable MDP over all ${k-1}$ rock combinations, which scales exponentially in $n$.
In problems with higher state space dimensions, like \textsc{RockSample}$(20,20)$, the QMDP solution is intractable.
Thus, fixed bounds are used in AdaOPS assuming an optimistic $V_\text{max}$ \citep{adaops2021review}.
The appendix further details the heuristics used by the baseline algorithms.
Indicated in \cref{tab:results}, the raw networks alone perform well but outperform when combined with online planning, enabling reasoning with current information.

\begin{figure}[t]
    \centering
    \includegraphics[height=3.5cm]{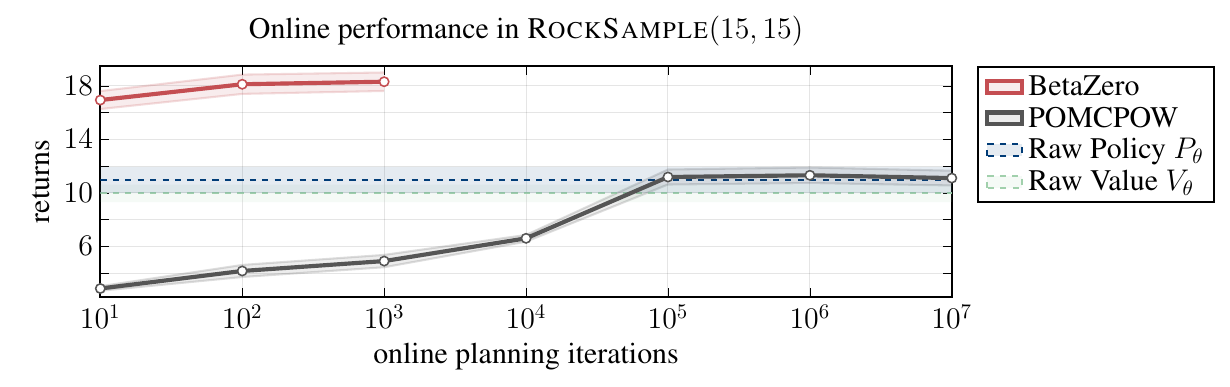}
    \caption{Performance of POMCPOW with heuristics up to $10$ million \textit{online} iterations plateaus, indicating that extending online searches alone misses valuable \textit{offline} experience.}
    \label{fig:online}
\end{figure}

\begin{figure*}[b!]
    \centering
    \resizebox{0.49\textwidth}{!}{%
        \includegraphics{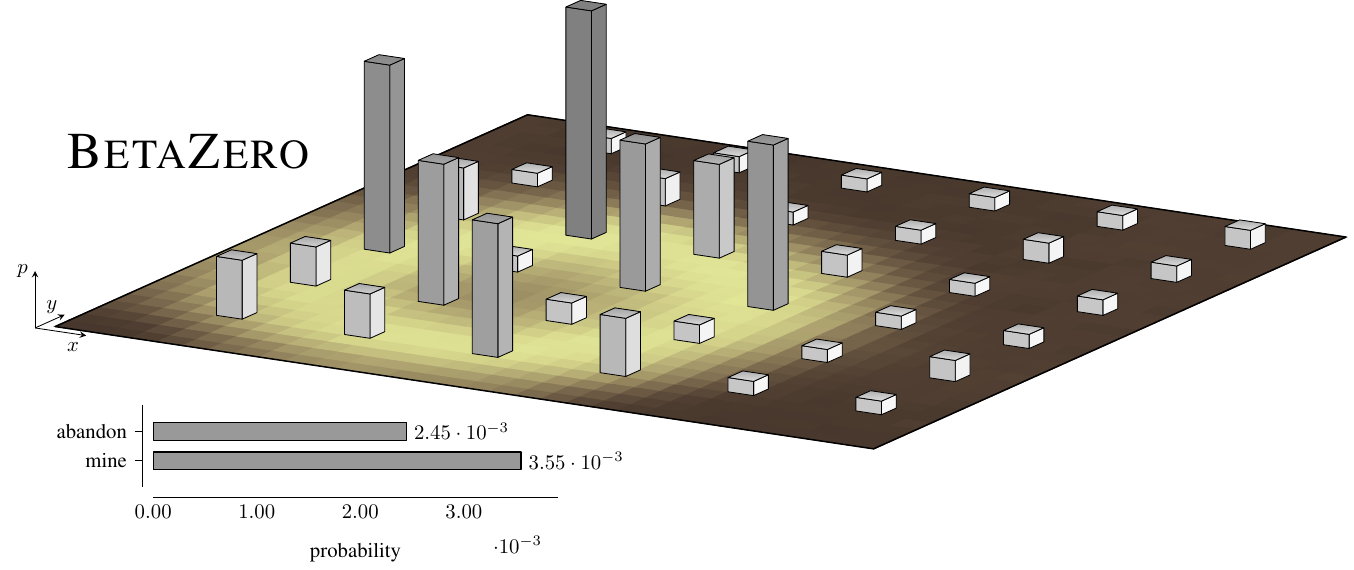}
    }
    \hfill
    \resizebox{0.49\textwidth}{!}{%
        \includegraphics{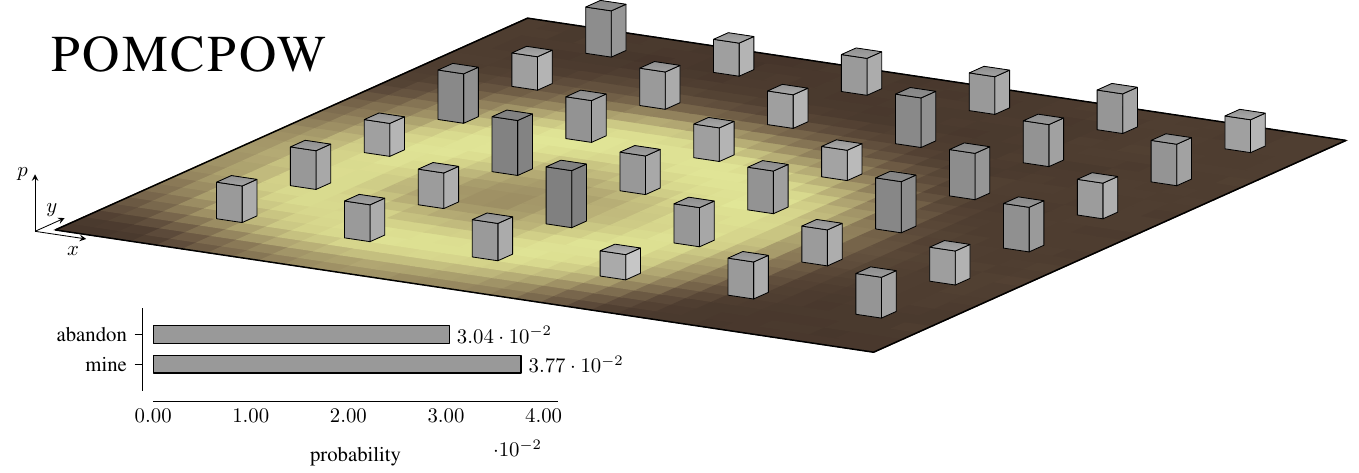}
    }
    \caption{Mineral exploration policies: BetaZero prioritizes uncertainty, matching heuristics from \citet{mern2023intelligent} (i.e., select action with high uncertainty, shown in yellow).}
    \label{fig:minex}
\end{figure*}

If online algorithms ran for a large number of iterations, one might expect to see convergence to the optimal policy.
In practice, this may be an intractable number as \cref{fig:online} shows POMCPOW has not reached the required number of iterations for RockSample.
The advantage of BetaZero is that it can generalize from a more diverse set of experiences.
The inability of existing online algorithms to plan over long horizons is also evident in the mineral exploration POMDP (\cref{fig:minex}).
POMCPOW ran for one million online iterations without a value estimator heuristic and BetaZero ran online for $100$ iterations (using about $850{,}000$ offline simulations).
In the figure, the probability of selecting a drilling location is shown as vertical bars for each action, overlaid on the initial belief uncertainty (i.e., the std of the belief in subsurface ore quality).
BetaZero learned to take actions in areas of the belief space with high uncertainty and high value (which matches the domain-specific heuristics developed for the mineral exploration problem from \citet{mern2023intelligent}), while POMCPOW fails to distinguish between the actions and resembles a uniform policy.

The most closely related algorithm, LeTS-Drive \citep{cai2022closing}, which also includes an offline learning component with online tree search planning, performs better than its DESPOT counterpart without the use of offline heuristics.
This is observed in all studied POMDPs except for the mineral exploration problem where the DESPOT bounds use privileged information from the approximately optimal bounds on the value function.
\Cref{tab:results} highlights that LeTS-Drive is able to scale DESPOT to the \textsc{RockSample}$(20,20)$ problem, with overall similar timing results as BetaZero but worse performance.
This could be attributed to the HyP-DESPOT online tree search used in LeTS-Drive that plans over observation space (similar to POMCPOW) and implicitly constructs beliefs from a set of $K$ scenarios in the tree.
Therefore, the beliefs are dependent on the number of in-tree scenarios executed, hence the comparable timing results, and not on the actual root node belief that is updated along the tree paths (where belief-state planning incurs different computational expense but with the benefit of planning over reachable beliefs into the future).
Instead of the state history as network input, we use the in-tree belief for a better comparison.
The HyP-DESPOT planner expands the tree over \textit{all} actions instead of using progressive widening with prioritization, and, as we observe in the ablation studies in the next section, expanding on all actions may limit the effective use of the tree search budget, thus potentially missing promising areas of the reachable futures.

\begin{figure*}[t!]
    \centering
    \begin{minipage}{0.69\textwidth}
        \resizebox{\textwidth}{!}{
            \includegraphics{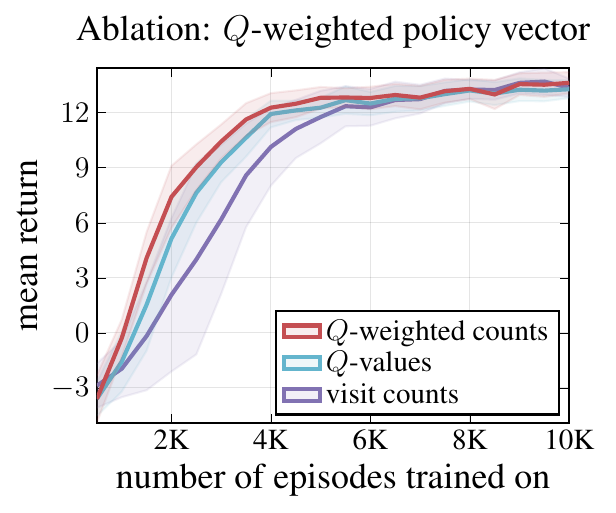}
            \includegraphics{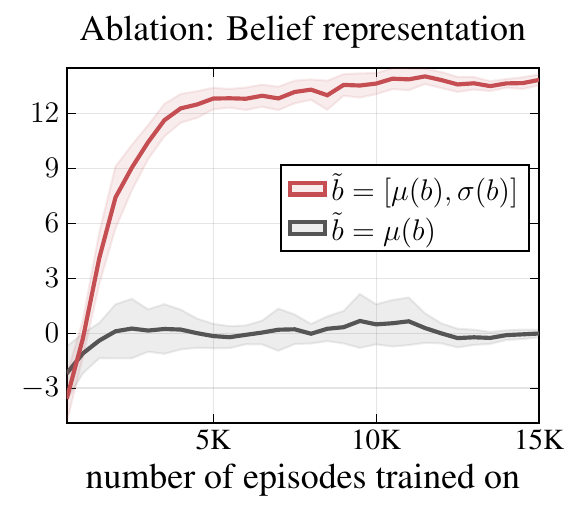}
            \includegraphics{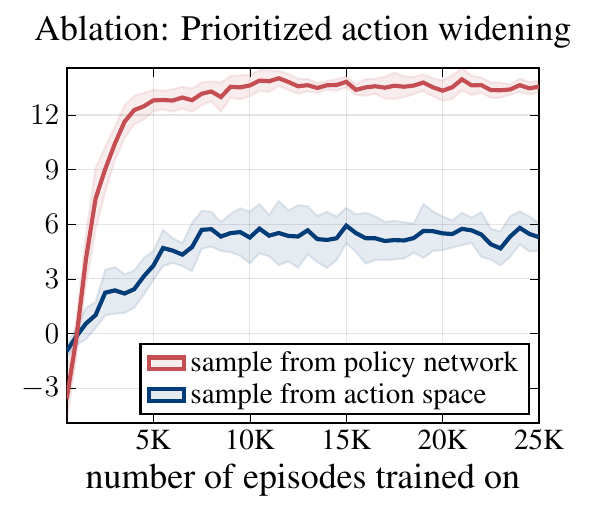}
    }
    \end{minipage}
    \hspace*{3mm} 
    \begin{minipage}{0.3\textwidth}
        \resizebox{\textwidth}{!}{%
            \includegraphics{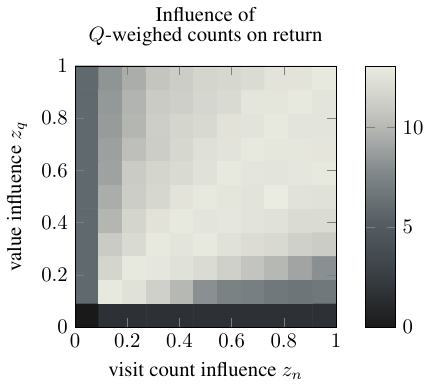}
        }
    \end{minipage}

    \begin{minipage}[t]{0.69\textwidth}
        \caption{
        \textsc{LightDark}$(10)$ ablation study.
        (Left) Learning is faster when the network is trained using $Q$-weighted visit counts.
        (Middle) Incorporating belief uncertainty is crucial for learning.
        (Right) Action widening from the policy network shows significant improvement.
        The same red curves are shown with varying horizontal axes, and one std is shaded from three seeds using $0.6$ exponential smoothing.}
        \label{fig:ablations}
    \end{minipage}
    \hspace*{3mm} 
    \begin{minipage}[t]{0.3\textwidth}
        \caption{Ablation study in \textsc{RockSample}$(20,20)$.
        Combining value and count information leads to the highest return. The diagonal is identical due to the $\argmax$ of \cref{eq:policy_q_weight}.}
        \label{fig:z_sweep}
    \end{minipage}
\end{figure*}

\paragraph{Ablation studies.}
To test the effect of each contribution, we run several ablation studies.
The influence of value and visit count information when selecting an action is shown in \cref{fig:z_sweep}.
Each cell is the mean return for the \textsc{RockSample}$(20,20)$ problem over $100$ online trials, selecting root-node actions via the $\argmax$ of \cref{eq:policy_q_weight} given $z_q$ and $z_n$.
The cell at $(0,0)$ corresponds to a uniform policy and thus samples actions instead.
Using only the visit counts (bottom cells) or only the values (left cells) to make decisions is worse than using a combination of the two.
The effect of the $Q$-weighting is also shown in the leftmost \cref{fig:ablations}, which suggests that it helps learn faster in \textsc{LightDark}$(10)$.

Unsurprisingly, using the \textit{state uncertainty} encoded in the belief is crucial for learning as indicated in the middle of \cref{fig:ablations}.
Future work could directly input the particle set into the network, first passing through an order invariant layer \citep{zaheer2017deep}, to offload the belief approximation to the network itself.
Finally, the rightmost plot in \cref{fig:ablations} suggests that when branching on actions using progressive widening, it is important to first prioritize the actions suggested by the policy network.
Offline learning fails if instead we sample uniformly from the action space (even in the \textsc{LightDark} case with only three actions).

\section{Conclusions}
We propose the \textit{BetaZero} belief-state planning algorithm for POMDPs; designed to learn from \textit{offline} experience to inform \textit{online} decisions.
Planning in belief space explicitly handles state uncertainty and learning offline approximations to replace heuristics enables effective online planning in long-horizon POMDPs.
Although belief-space planning incurs expensive belief updates in the tree search, we address the limited search budget used in practice by incorporating all information available in the search tree to (a) train the policy vector target (using the $Q$-weighted visit counts), and (b) sample from the policy network during action progressive widening to prioritize promising actions.
Stochastic belief-state transitions in MCTS are addressed using \textit{progressive widening} and we test a belief representation of summary statistics to allow beliefs as input to the value and policy network.
Results indicate that BetaZero scales to larger problems where certain heuristics break down and, as a result, can solve large-scale POMDPs by learning to plan in belief space using zero heuristics.

\paragraph{Limitations.}
It is standard for POMDP planning algorithms to assume known models but this may limit the applicability to certain problems where reinforcement learning may be better suited.
We chose a simplified belief representation to allow for further research innovations in using other parametric and non-parametric representations.
Other limitations include compute resource requirements for training neural networks and parallelizing MCTS simulations.
We designed BetaZero to use a single GPU for training and to scale based on available CPUs.
Certain POMDPs may not require this training burden, especially when known heuristics perform well.
BetaZero is useful for long-horizon, high-dimensional POMDPs but may be unnecessary when offline training is computationally limited.
BetaZero is designed for problems where the simulation cost is the dominating factor compared to offline training time.

\section*{Acknowledgments}
We would like to thank Johannes Fischer, Arec Jamgochian, Markus Zechner, Dylan Asmar and Ashwin Kanhere for their insights and comments on this work.
This work is supported by funding from the Stanford Institute for Human-Centered AI (HAI), Stanford Mineral-X, and OMV.

\bibliography{references}
\bibliographystyle{preamble/rlc}

\clearpage
\appendix

\section*{Appendix}

This section contains material detailing the POMDP environments and experiments, the ablation studies, additional analysis of bootstrapping and double progressive widening, the network architectures, hyperparameters and tuning, computational resources, information regarding open-source code for reproducibility, and the full BetaZero algorithm pseudocode.

\section{POMDP Environments}
This section describes the benchmark POMDPs in detail, including the heuristics used by the baseline POMDP algorithms and information regarding the particle filter belief used by BetaZero.

\paragraph{Light dark.}

\begin{figure}[t!]
    \centering
    \includegraphics[width=0.75\linewidth]{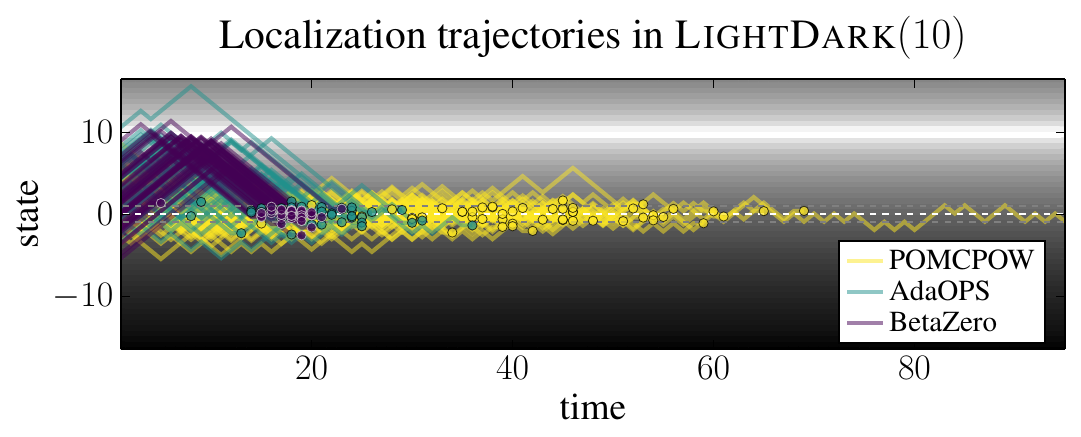}
    \caption{$\textsc{LightDark}(10)$ trajectories from $50$ episodes.
    BetaZero (dark blue) learned to first localize in the light region at $y=10$ before heading to the goal (origin).}
    \label{fig:lightdark_trajectories}
\end{figure}

The \textsc{LightDark}$(y)$ POMDP is a one-dimensional localization problem \citep{platt2010belief}.
The objective is for the agent to execute the \textit{stop} action at the goal, which is at $\pm 1$ of the origin.
The agent is awarded $100$ for stopping at the goal and $-100$ for stopping anywhere else; using a discount of $\gamma = 0.9$.
The agent receives noisy observations of their position, where the noise is minimized in the ``light'' region defined by $y$.
In the $\textsc{LightDark}(5)$ problem used by \citet{wu2021adaptive}, the noise is a zero-mean Gaussian with standard deviation of $|y - 5|/\sqrt{2} + 10^{-2}$.
For the $\textsc{LightDark}(10)$ problem used by \citet{sunberg2018online}, the noise is a zero-mean Gaussian with standard deviation of $|y - 10| + 10^{-4}$.
In both problems, we use a restricted action space of $\mathcal{A} = [-1, 0, 1]$ where $0$ is the \textit{stop} action.
The expected behavior of the optimal policy is first to localize in the light region, then travel down to the goal.
The BetaZero policy exhibits this behavior which can be seen in \cref{fig:lightdark_trajectories} (where circles indicate the final location).

The approximately optimal solution to the light dark problems used \textit{local approximation value iteration} (LAVI) \citep{dmubook} over the discretized belief-state space (i.e., mean and std).
The belief mean was discretized between the range $[-12, 12]$ and the belief std was discretized between the range $[0, 5]$; each of length $100$.
The LAVI solver used $100$ generative samples per belief state and ran for $100$ value iterations with a Bellman residual of $\num{1e-3}$.

\paragraph{Rock sample.}

In the $\textsc{RockSample}(n,k)$ POMDP introduced by \citet{smith2012heuristic}, an agent has full observability of its position on an $n \times n$ grid but has to sense the $k$ rocks to determine if they are ``good'' or ``bad''.
The agent knows \textit{a priori} the true locations of the rocks (i.e., the rock locations $\mathbf{x}_\text{rock}$ are a part of the problem, not the state).
The observation noise is a function of the distance to the rock:
\begin{equation}
   \frac{1}{2}\left(1 + \exp\left(-\frac{\lVert \mathbf{x}_\text{rock} - \mathbf{x}_\text{agent} \rVert_2 \log(2)}{c}\right)\right)
\end{equation}
where $c=20$ is the sensor efficiency.
The agent can move in the four cardinal directions, sense the $k$ rocks, or take the action to \textit{sample} a rock when it is located under the agent.
The agent receives a reward of $10$ for sampling a ``good'' rock and a penalty of $-10$ for sampling a ``bad'' rock.
The terminal state is the exit at the right edge of the map, where the agent gets a reward of $10$ for exiting.

\paragraph{Mineral exploration.}

\begin{figure*}[hp!]
    \centering
    \begin{minipage}{0.65\textwidth}
        \resizebox{\textwidth}{!}{%
            \includegraphics{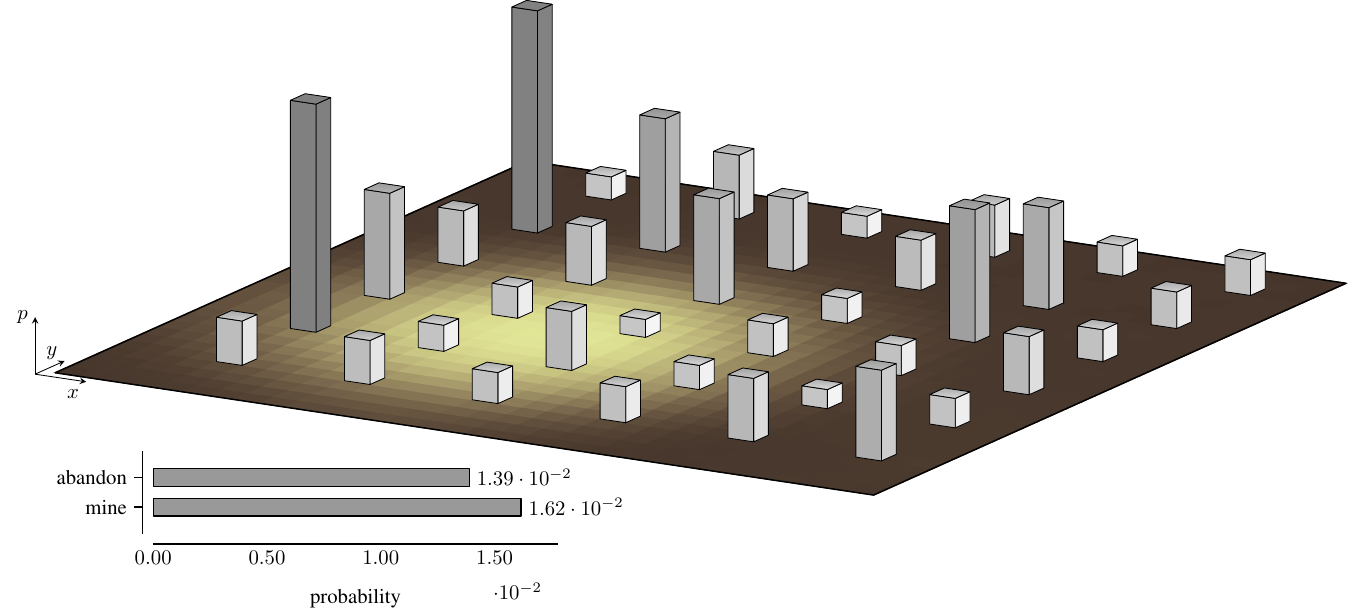}
        }
    \end{minipage}
    \hfill
    \begin{minipage}{0.3\textwidth}
        \resizebox{\textwidth}{!}{%
            \includegraphics{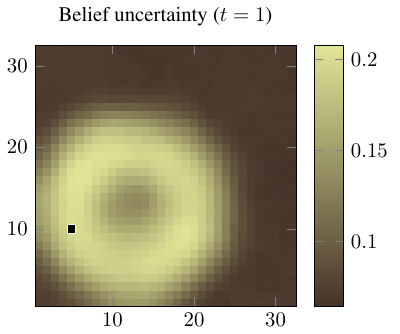}
        }
    \end{minipage}
    \begin{minipage}{0.65\textwidth}
        \resizebox{\textwidth}{!}{%
            \includegraphics{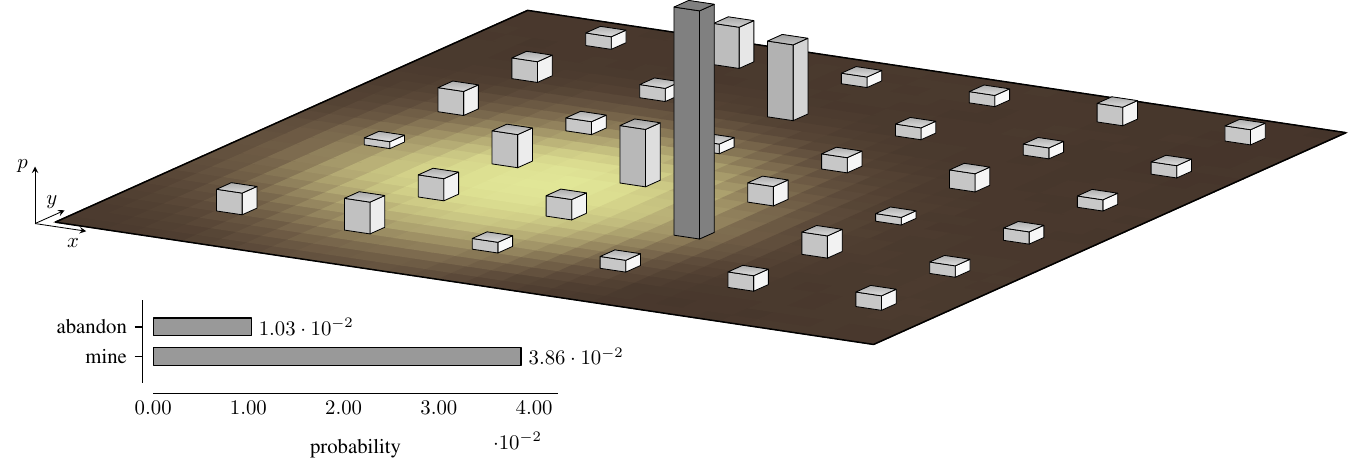}
        }
    \end{minipage}
    \hfill
    \begin{minipage}{0.3\textwidth}
        \resizebox{\textwidth}{!}{%
            \includegraphics{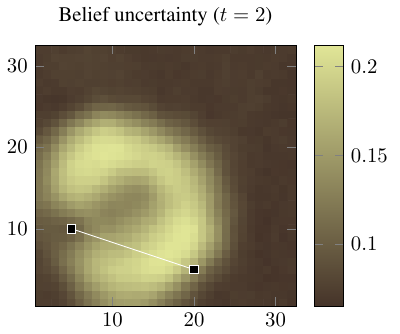}
        }
    \end{minipage}
    \begin{minipage}{0.65\textwidth}
        \resizebox{\textwidth}{!}{%
            \includegraphics{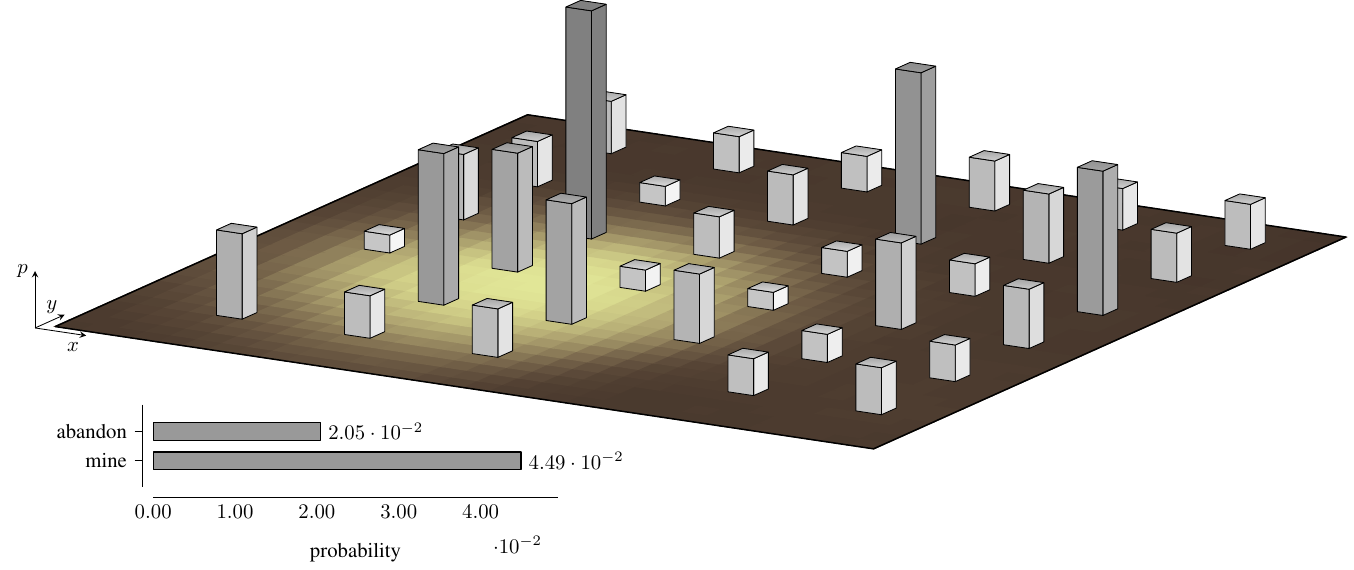}
        }
    \end{minipage}
    \hfill
    \begin{minipage}{0.3\textwidth}
        \resizebox{\textwidth}{!}{%
            \includegraphics{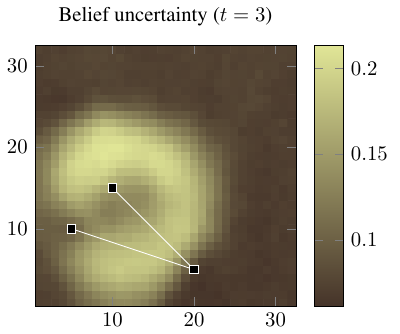}
        }
    \end{minipage}
    \begin{minipage}{0.65\textwidth}
        \resizebox{\textwidth}{!}{%
            \includegraphics{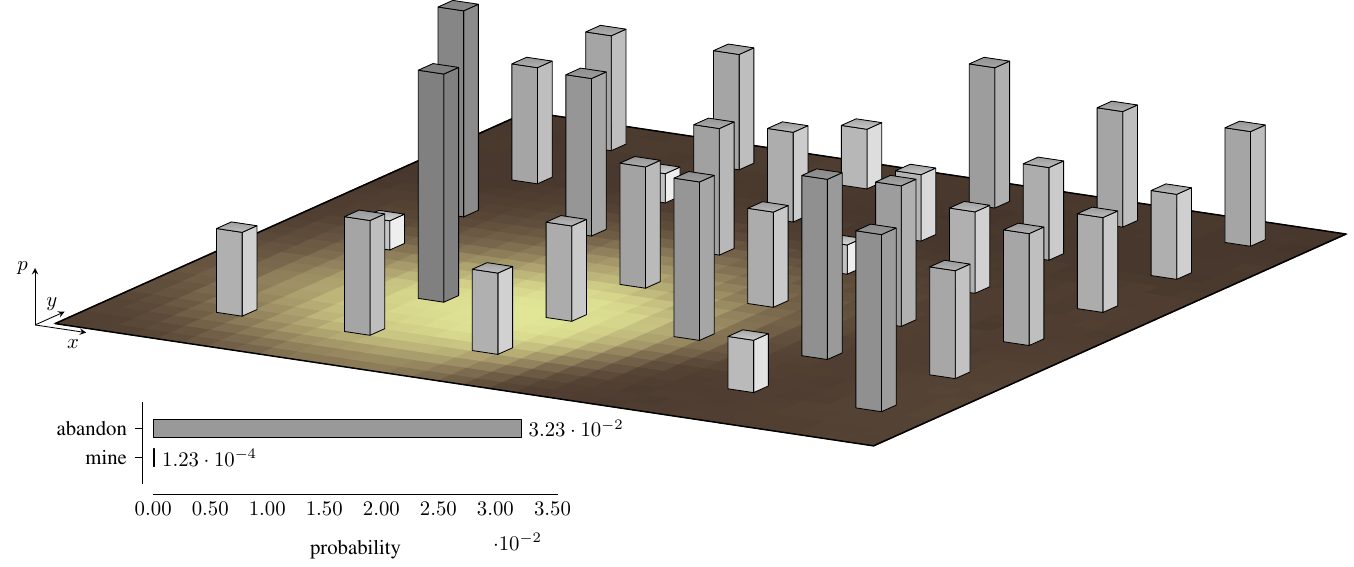}
        }
    \end{minipage}
    \hfill
    \begin{minipage}{0.3\textwidth}
        \resizebox{\textwidth}{!}{%
            \includegraphics{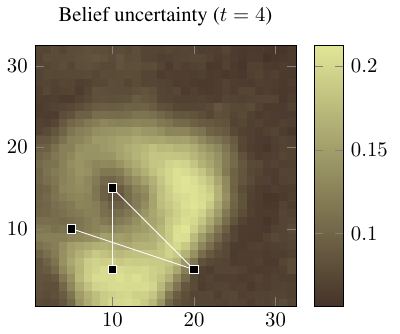}
        }
    \end{minipage}

    \begin{minipage}[t]{0.65\textwidth}
        \caption{The BetaZero policy shown over belief mean for four steps. BetaZero first prioritizes the edges of the belief mean, corresponding to the belief uncertainty (right-most plots), then explores the outer regions of the subsurface; ultimately gathering information from actions with high mean and std, matching heuristics. At the initial step, abandoning and mining have near-equal probability (bottom left graphs) but by the fourth action, abandoning is much more likely.}
        \label{fig:minex_policy}
    \end{minipage}
    \hfill
    \begin{minipage}[t]{0.3\textwidth}
        \caption{The selected drill actions over belief uncertainty, showing that uncertainty collapses after drilling.}
        \label{fig:minex_2d}
    \end{minipage}
\end{figure*}

The \textsc{Mineral Exploration} POMDP introduced by \citet{mern2023intelligent} is an information gather problem with the goal of deciding whether a subsurface ore body is economical to mine or should be abandoned (calibrated so that $50\%$ of cases are economical).
The agent can drill every fifth cell of a $32 \times 32$ plot of land to determine the ore quality at that location.
Therefore, the action space consists of the $36$ drill locations and the final decisions to either \textit{mine} or \textit{abandon}.
The agent receives a small cost for each \textit{drill} action, a reward proportional to the extracted ore if chosen to \textit{mine} (which is negative if uneconomical), and a reward of zero if chosen to \textit{abandon}:
\begin{equation}
R(s,a) = \begin{cases}
    -c_\text{drill} & \text{if } a=\text{drill}\\
    \sum\mathds{1}(s_\text{ore} \ge h_\text{massive}) - c_\text{extract} & \text{if } a=\text{mine}\\
    0 & \text{otherwise}
\end{cases}\label{eq:minex_reward}
\end{equation}
where $c_\text{drill}=0.1$, $h_\text{massive}=0.7$, and $c_\text{extract}=71$.
The term $\sum\mathds{1}(s_\text{ore} \ge h_\text{massive})$ indicates the cells that have an ore quality value above some massive ore threshold $h_\text{massive}$ (which are deemed valuable).
\Cref{fig:minex_policy} and \cref{fig:minex_2d} show an example of four steps of the mineral exploration POMDP.

\subsection{Experiment details}

Experiment parameters for each problem can be seen in \cref{tab:ld_params,tab:rs_params,tab:minex_params} under the ``online'' column.
For the baseline algorithms, the heuristics follow \citet{wu2021adaptive}.
Problems that failed to run due to memory limits followed suggestions from \citet{adaops2021review} to first use the MDP solution and then use a fixed upper bound of $r_\text{correct}=100$ for the light dark problems and the following for the rock sample problems:
\begin{equation}
    V_\text{max} =  r_\text{exit} + \sum_{t=1+n-k}^{2k-n} \gamma^{t-1}r_\text{good}
\end{equation}
where $r_\text{good} = r_\text{exit} = 10$ and the sum computes an optimistic value assuming the rocks are directly lined between the agent and the goal and assuming $n \ge k$ for simplicity.

For problems not studied by \citet{wu2021adaptive}, we use the same heuristics as their easier counterpart (i.e., $\textsc{LightDark}(10)$ uses $\textsc{LightDark}(5)$ heuristics and $\textsc{RockSample}(20,20)$ uses $\textsc{RockSample}(15,15)$ heuristics).
For mineral exploration, the baselines used the following heuristics.
POMCPOW used a value estimator of $\max(0, R(s, a=\text{mine}))$ and when using ``no heuristic'' used a random rollout policy to estimate the value.
Both AdaOPS and DESPOT used a lower bound computed as the returns if fully drilled all locations, then made the decision to abandon:
\begin{equation}
    V_\text{min} = -\sum_{t=1}^{T-1} \gamma^{t-1}c_\text{drill}
\end{equation}
The upper bound comes from an oracle $\pi_\text{truth}$ taking the correct final action without drilling, computed over $10{,}000$ states. Note that there is no state transition in this problem.
\begin{align}
    V_\text{max} &= \operatorname*{\mathbb{E}}_{s \in \mathcal{S}} \bigg[ \max\Big(0, R\big(s, \pi_\text{truth}(s)\big)\Big) \bigg] \\
                 &\approx \frac{1}{n}\sum_{i=1}^n \max\Big(0, R\big(s^{(i)}, \pi_\text{truth}(s)\big)\Big)
\end{align}

\paragraph{Particle filtering.}
Both BetaZero and the baseline algorithms update their belief with a bootstrap particle filter using a low-variance resampler \citep{gordon1993novel}, with $n_\text{particles} \in [500,1000,1000]$ for the light dark, rock sample, and mineral exploration problems, respectively.
The particle filter follows an update procedure of first reweighting then resampling.
In mineral exploration, the observations are noiseless which could quickly result in particle depletion.
Therefore, approximate Bayesian computation (ABC) is used to reweight each particle using a Gaussian distribution centered at the observation with a standard deviation of $\sigma_\text{abc} = 0.1$ \citep{csillery2010approximate}.

The belief representation takes the mean and standard deviation across the $n_\text{particles}$. In the light dark problems, this is computed across the $500$ sampled $y$-state values that make up the belief.
The initial $y$-value state distribution---which makes up the initial belief---follows a Gaussian distribution and thus the parametric representation is a good approximation of the belief.

For the rock sample problem, the belief is represented as the mean and standard deviation of the good rocks from the $1000$ sampled states (appending the true position as it is deterministic).
The rock qualities are sampled uniformly in $\{0,1\}$ indicating if they are ``good'', which makes the problem non-Gaussian, but the parametric belief approximation can model a uniform distribution by placing the mean at the center of the uniform range and stretching the variance to match the uniform.

Lastly, the mineral exploration problem flattens the $1000$ subsurface $32\times32$ maps that each have associated ore quality per-pixel between $[0,1]$ into two images: a mean and standard deviation image of the ore quality that is stacked and used as input to a CNN.
The initial state distribution for the massive ore quantity closely follows a Gaussian, making the parametric belief approximation well suited.

For problems where Gaussian approximations do not capture the belief, the parameters of other distributions could be used as a belief representation or the particles themselves could be input into a network---first passing through an order-invariant layer \citep{igl2018deep}.
Scaling to larger observation spaces will not be an issue as BetaZero plans over belief states instead of observations.

\section{Additional Analysis}

This section briefly describes additional analyses omitted from the main body of the paper.
This includes analysis of bootstrapping the initial $Q$-values using a one-step lookahead with the value network and sensitivity analysis of double progressive widening on belief-states and actions.

\subsection{Bootstrapping analysis}

When adding a belief-action pair $(b,a)$ to the MCTS tree, initializing the $Q$-values via bootstrapping with the value network may improve performance when using a small MCTS budget.
\Cref{tab:bootstrap} shows the results of an analysis comparing BetaZero with bootstrapping $Q_0(b,a) = R_b(b,a) + \gamma V_\theta(\tilde{b}')$ where $\tilde{b}' = \phi(b')$ and without bootstrapping $Q_0(b,a)=0$.
Each domain used the online parameters described in \cref{tab:ld_params,tab:rs_params,tab:minex_params}.
Results indicate that bootstrapping was only helpful in the rock sample problems and incurs additional compute time due to the belief update done in $b' \sim T_b(b,a)$.
Note that bootstrapping was not used during offline training.
In problems with high stochasticity in the belief-state transitions, bootstrapping may be noisy during the initial search due to the transition $T_b$ sampling a single state from the belief.
Further analysis could investigate the use of multiple belief transitions to better estimate the value, at the expense of additional computation.
The value estimate of $b$ could instead be used as the bootstrap but we would expect similar results to the one-step bootstrap as many problems we study have sparse rewards.

\begin{table*}[b]
    \centering
    \begin{threeparttable}
        \begin{adjustbox}{max width=\textwidth}
        \begin{tabular}{@{}lrrrrrrrrrr@{}}
            \arrayrulecolor{black} 
            \toprule
                & \multicolumn{2}{c}{$\text{LightDark}(5)$}  &  \multicolumn{2}{c}{$\text{LightDark}(10)$}  &  \multicolumn{2}{c}{$\text{RockSample}(15,15)$}  &  \multicolumn{2}{c}{$\text{RockSample}({20,20})$}  &  \multicolumn{2}{c}{Mineral Exploration} \\
            \arrayrulecolor{lightgray}
            \cmidrule{2-11}
            \arrayrulecolor{black} 
                & returns & time [s] & returns & time [s] & returns & time [s] & returns & time [s] & returns & time [s] \\
            \midrule
            \arrayrulecolor{white}
            BetaZero (bootstrap)     &  $\num{4.22 \pm 0.31}$  &  $\num{0.014}$  &  $\num{14.45 \pm 1.15}$  &  $\num{0.34}$  &  $\mathBF{20.15 \pm 0.71}$  &  $\num{0.48}$  &  $\mathBF{13.09 \pm 0.55}$  &  $\num{1.11}$  &  $\num{10.32 \pm 2.38}$  &  $\num{6.27}$  \\
            \midrule
            BetaZero (no bootstrap)  &  $\mathBF{4.47 \pm 0.28}$  &  $\num{0.014}$  &  $\mathBF{16.77 \pm 1.28}$  &  $\num{0.33}$  &  $\num{19.50 \pm 0.71}$  &  $\num{0.42}$  &  $\num{11.00 \pm 0.54}$  &  $\num{0.57}$  &  $\mathBF{10.67 \pm 2.25}$  &  $\num{4.46}$  \\
            \arrayrulecolor{black} 
            \bottomrule
        \end{tabular}
        \end{adjustbox}
        \begin{tablenotes}
            \small
            \item[\phantom{*}] {Reporting mean $\pm$ standard error over $100$ seeds (i.e., episodes); timing is average per episode.}
        \end{tablenotes}
    \end{threeparttable}
    \caption{Effect of $Q$-value bootstrapping in online \textit{BetaZero} performance (returns and online timing).}\label{tab:bootstrap}
\end{table*}

\subsection{Limitations of double progressive widening}

Double progressive widening (DPW) is a straightforward approach to handle large or continuous state and action spaces in Monte Carlo tree search.
It is easy to implement and only requires information available in the tree search, i.e., number of children nodes and number of node visits.
It is known that MCTS performance can be sensitive to DPW hyperparameter tuning and \citet{sokota2021monte} show that DPW ignores information about the relation between states that could provide more intelligent branching.
\citet{sokota2021monte} introduce \textit{state abstraction refinement} that uses a distance metric between states to determine if a similar state should be added to the tree; requiring a state transition every time a state-action node is visited.
For our work, we want to reduce the number of expensive belief-state transitions in the tree and avoid the use of problem-specific heuristics required when defining distance metrics.
Using DPW in BetaZero was motivated by simplicity and allows future work to innovate on the components of belief-state and action branching.

\begin{figure}[t!]
    \captionsetup{font={small}}
    \centering
    \begin{minipage}{0.24\linewidth}
        \resizebox{\linewidth}{!}{
            \includegraphics{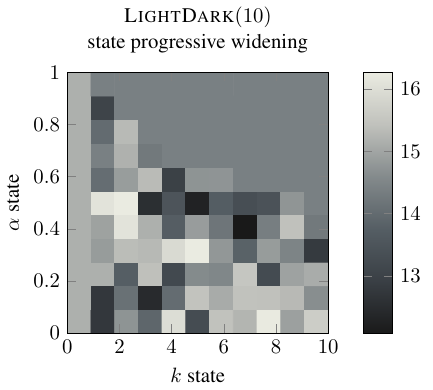}
    }
    \end{minipage}
    \hfill
    \begin{minipage}{0.24\linewidth}
        \resizebox{\linewidth}{!}{%
            \includegraphics{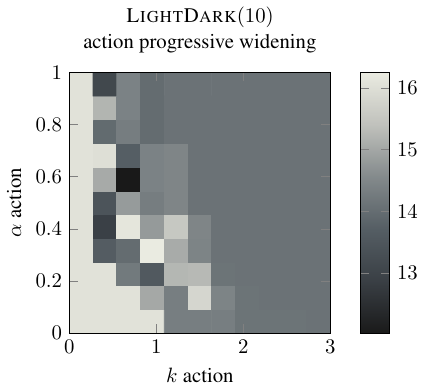}
        }
    \end{minipage}
    \hfill
    \begin{minipage}{0.24\linewidth}
        \resizebox{\linewidth}{!}{
            \includegraphics{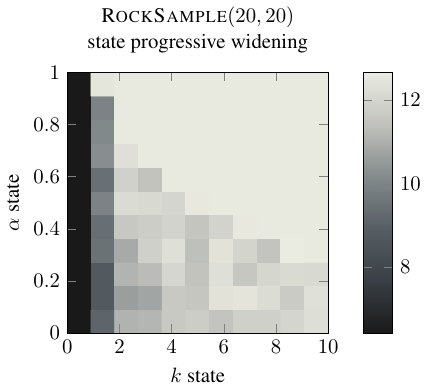}
    }
    \end{minipage}
    \hfill
    \begin{minipage}{0.24\linewidth}
        \resizebox{\linewidth}{!}{%
            \includegraphics{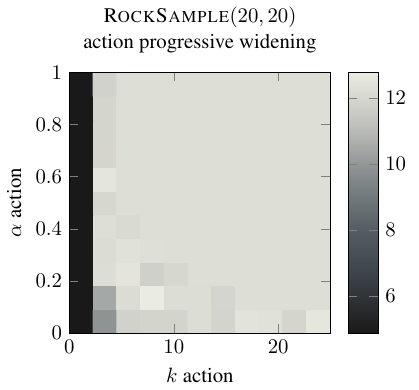}
        }
    \end{minipage}
    \begin{minipage}[t]{0.23\linewidth}
        \caption{Sensitivity analysis of \textit{belief-state} progressive widening in $\textsc{LightDark}(10)$.}
        \label{fig:spw_sensitivity_ld10}
    \end{minipage}
    \hfill
    \begin{minipage}[t]{0.23\linewidth}
        \caption{Sensitivity analysis of \textit{action} progressive widening in $\textsc{LightDark}(10)$.}
        \label{fig:apw_sensitivity_ld10}
    \end{minipage}
    \hfill
    \begin{minipage}[t]{0.23\linewidth}
        \caption{Sensitivity analysis of \textit{belief-state} progressive widening in $\textsc{RockSample}(20,20)$.}
        \label{fig:spw_sensitivity_rs20}
    \end{minipage}
    \hfill
    \begin{minipage}[t]{0.23\linewidth}
        \caption{Sensitivity analysis of \textit{action} progressive widening in $\textsc{RockSample}(20,20)$.}
        \label{fig:apw_sensitivity_rs20}
    \end{minipage}
\end{figure}

To analyze the sensitivity of DPW, \cref{fig:spw_sensitivity_ld10,fig:apw_sensitivity_ld10} show a sweep over the $\alpha$ and $k$ parameters for DPW in $\textsc{LightDark}(10)$.
\Cref{fig:spw_sensitivity_ld10} shows that the light dark problem is sensitive to belief-state widening and \cref{fig:apw_sensitivity_ld10} indicates that this problem may not require widening on all actions---noting that when $k=0$, the only action expanded on is the one prioritized from the policy head $a\sim P_\theta(\tilde{b}, \cdot)$.
The light dark problems have a small action space of ${|\mathcal{A}|=3}$, therefore this prioritization leads to good performance when only a single action is evaluated (left cells in \cref{fig:apw_sensitivity_ld10} when $k=0$).

In $\textsc{RockSample}(20,20)$, \cref{fig:spw_sensitivity_rs20,fig:apw_sensitivity_rs20} indicates that this problem benefits from a higher widening factor (top right of the figures) as the action space ${|\mathcal{A}|=25}$ is larger and the belief-state transitions operate over a much larger state space.
DPW uses a single branching factor throughout the tree search and research into methods that adapt the branching based on learned information would be a valuable direction to explore.

\citet{lim2023optimality} introduce a class of POMDP planning algorithms that use a fixed number of samples to branch on instead of progressive widening.
The bottom row of \cref{fig:spw_sensitivity_ld10,fig:apw_sensitivity_ld10,fig:spw_sensitivity_rs20,fig:apw_sensitivity_rs20} (where $\alpha = 0$) can be interpreted as a fixed branching factor compared to progressive widening in the other cells. The analysis in the figures shows that there are cases where BetaZero has better performance when using progressive widening (show in the lighter colors).

\begin{figure*}[ht!]
    \centering
    \begin{minipage}{0.22\textwidth}
        \resizebox{\textwidth}{!}{%
            \includegraphics{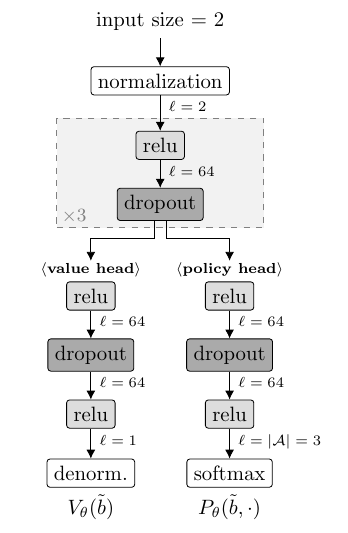}
        }
    \end{minipage}
    \hfill
    \begin{minipage}{0.28\textwidth}
        \resizebox{\textwidth}{!}{%
            \includegraphics{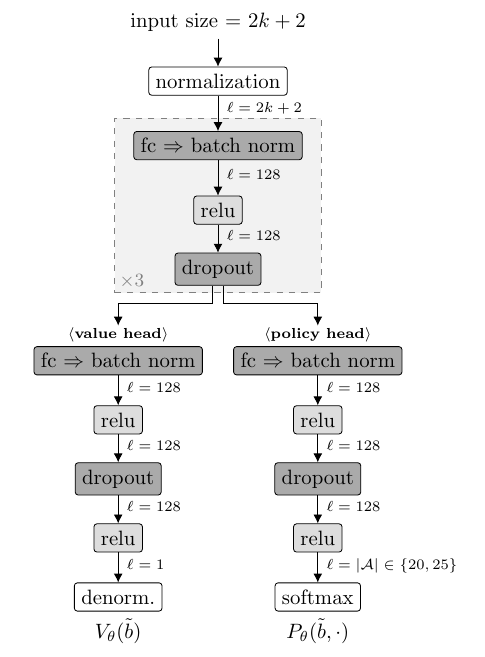}
        }
    \end{minipage}
    \hfill
    \begin{minipage}{0.26\textwidth}
        \resizebox{\textwidth}{!}{%
            \includegraphics{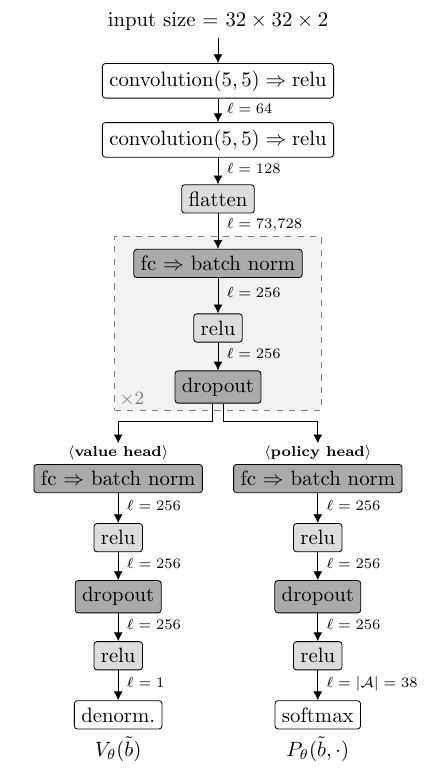}
        }
    \end{minipage}

    \begin{minipage}[t]{0.22\textwidth}
        \caption{Light dark neural network architecture.}
        \label{fig:nn_lightdark}
    \end{minipage}
    \hfill
    \begin{minipage}[t]{0.28\textwidth}
        \caption{Rock sample neural network architecture.}
        \label{fig:nn_rocksample}
    \end{minipage}
    \hfill
    \begin{minipage}[t]{0.26\textwidth}
        \caption{Mineral exploration CNN architecture.}
        \label{fig:nn_minex}
    \end{minipage}
\end{figure*}

\section{Network Architectures}

\Cref{fig:nn_lightdark,fig:nn_rocksample,fig:nn_minex} specify the neural network architectures for the three problem domains.
The networks were designed to be simple so that future work could focus on incorporating more complicated architectures such as residual networks.
Mineral exploration does not normalize the inputs and is the only problem where the input is treated as an image, thus we use a convolutional neural network (CNN).
Training occurs on normalized returns and an output denormalization layer is added to the value head to ensure proper magnitude of the predicted values.

\subsection{Return scaling for output normalization}

For general POMDPs, the return can be an unbounded real-value and not conveniently in $[0,1]$ or $[-1,1]$; as is often the case with two player games. 
\citet{schrittwieser2020mastering} use a categorical representation of the value split into a discrete support to make learning more robust \citep{schrittwieser2020intuition}.
We instead simply normalize the target before training as
\begin{equation}
    \bar{g}_t = \frac{g_t - \mathbb{E}[G_\text{train}]}{\sqrt{\Var[G_\text{train}]}}
\end{equation}
where $G_\text{train}$ is the set of returns used during training; keeping running statistics of all training data.
Intuitively, this ensures that the target values have zero mean and unit variance which is known to stabilize training \citep{lecun2002efficient}.
After training, a denormalization layer is added to the normalized output $\bar{v}$ of the value network as
\begin{equation}
    v_t = \bar{v} \sqrt{\Var[G_\text{train}]} + \mathbb{E}[G_\text{train}]
\end{equation}
to properly scale value predictions when the network is evaluated (which is done entirely internal to the network).

\section{Hyperparameters and Tuning}

The hyperparameters used during offline training and online execution are described in \cref{tab:ld_params,tab:rs_params,tab:minex_params}.
Offline training refers to the BetaZero policy iteration steps that collect parallel MCTS data (\textit{policy evaluation}) and then retrain the network (\textit{policy improvement}).
The online execution refers to using the BetaZero policy after offline training to evaluate its performance through online tree search.
The main difference between these two settings is the final criteria used to select the root node action in MCTS.
During offline training of problems with large action spaces (e.g., rock sample and mineral exploration), sampling root node actions according to the $Q$-weighted visit counts with a temperature $\tau$ ensures exploration.
To evaluate the performance online, root node action selection takes the maximizing action of the $Q$-weighted visit counts.
During training, we also evaluate a holdout set that uses the $\argmax$ criteria to monitor the true performance of the learned policy.

\begin{table*}[pt!]
    \centering
    \begin{adjustbox}{max width=\textwidth}
    \begin{threeparttable}
        \begin{footnotesize}
        \begin{tabular}{@{}clrrrrm{9cm}@{}}
            \arrayrulecolor{black} 
            \toprule
             & \multirow{2}{*}{Parameter\tnote{*}}  &  \multicolumn{2}{c}{$\text{LightDark}(5)$}  &  \multicolumn{2}{c}{$\text{LightDark}(10)$}  &  \multirow{2}{*}{Description} \\
            \arrayrulecolor{lightgray}
            \cmidrule{3-6}
            \arrayrulecolor{black} 
                    &  & Offline & Online & Offline & Online & \\
            \midrule
            \multirow{3}{*}{\makecell{BetaZero policy\\iteration parameters}} & $n_\text{iterations}$ & \num{30} & --- & \num{30} & --- & Number of offline BetaZero policy iterations. \\
             & $n_\text{data}$ & \num{500} & --- & \num{500} & --- & Number of parallel MCTS data gen. episodes per policy iteration. \\
             & bootstrap $Q_0$ & false & false & false & false & Use bootstrap estimate for initial $Q$-value in MCTS. \\
            \arrayrulecolor{lightgray}
            \midrule
            \multirow{3}{*}{\shortstack{Neural network\\parameters}} & $n_\text{epochs}$ & \num{50} & --- & \num{50} & --- & Number of training epochs. \\
             & $\alpha$ & \num{1e-4} & --- & \num{1e-4} & --- & Learning rate. \\
             & $\lambda$ & \num{1e-5} & --- & \num{1e-5} & --- & $L_2$-regularization parameter. \\
             \midrule
             \multirow{10}{*}{\shortstack{MCTS\\parameters}} & $n_\text{online}$ & \num{100} & \num{1300} & \num{100} & \num{1000} & Number of tree search iterations of MCTS.\\
              & $c$ & \num{1} & \num{1} & \num{1} & \num{1} & PUCT exploration constant. \\
              & $k_a$ & \num{2.0} & \num{2.0} & \num{2.0} & \num{2.0} & Multiplicative action progressive widening value.\\
              & $\alpha_a$ & \num{0.25} & \num{0.25} & \num{0.25} & \num{0.25} & Exponential action progressive widening value.\\
              & $k_b$ & \num{2.0} & \num{2.0} & \num{2.0} & \num{2.0} & Multiplicative belief-state progressive widening value.\\
              & $\alpha_b$ & \num{0.1} & \num{0.1} & \num{0.1} & \num{0.1} & Exponential belief-state progressive widening value.\\
              & $d$ & \num{10} & \num{10} & \num{10} & \num{10} & Maximum tree depth. \\
              & $\tau$ & \num{0} & \num{0} & \num{0} & \num{0} & Exploration temperature for final root node action selection. \\
              & $z_q$ & \num{1} & \num{1} & \num{1} & \num{1} & Influence of $Q$-values in final criteria. \\
              & $z_n$ & \num{1} & \num{1} & \num{1} & \num{1} & Influence of visit counts in final criteria. \\
            \arrayrulecolor{black} 
            \bottomrule
        \end{tabular}
        \begin{tablenotes}
            \item[*] {Entries with ``---'' denote non-applicability and ``$\cdot$'' denotes they are disabled.}
        \end{tablenotes}
        \end{footnotesize}
    \end{threeparttable}
    \end{adjustbox}
    \caption{\textit{BetaZero} parameters for the $\textsc{LightDark}$ problems.}\label{tab:ld_params}
\end{table*}

\begin{table*}[hp!]
    \centering
    \begin{adjustbox}{max width=\textwidth}
    \begin{threeparttable}
        \begin{footnotesize}
        \begin{tabular}{@{}clrrrrm{9cm}@{}}
            \arrayrulecolor{black} 
            \toprule
             & \multirow{2}{*}{Parameter}  &  \multicolumn{2}{c}{$\text{RockSample}(15,15)$}  &  \multicolumn{2}{c}{$\text{RockSample}(20,20)$}  &  \multirow{2}{*}{Description} \\
            \arrayrulecolor{lightgray}
            \cmidrule{3-6}
            \arrayrulecolor{black} 
                    &  & Offline & Online & Offline & Online & \\
            \midrule
            \multirow{3}{*}{\shortstack{BetaZero policy\\iteration parameters}} & $n_\text{iterations}$ & \num{50} & --- & \num{50} & --- & Number of offline BetaZero policy iterations. \\
             & $n_\text{data}$ & \num{500} & --- & \num{500} & --- & Number of parallel MCTS data gen. episodes per policy iteration. \\
             & bootstrap $Q_0$ & false & true & false & true & Use bootstrap estimate for initial $Q$-value in MCTS. \\
            \arrayrulecolor{lightgray}
            \midrule
            \multirow{3}{*}{\shortstack{Neural network\\parameters}} & $n_\text{epochs}$ & \num{10} & --- & \num{10} & --- & Number of training epochs. \\
             & $\alpha$ & \num{1e-3} & --- & \num{1e-3} & --- & Learning rate. \\
             & $\lambda$ & \num{1e-5} & --- & \num{1e-5} & --- & $L_2$-regularization parameter. \\
             \midrule
             \multirow{10}{*}{\shortstack{MCTS\\parameters}} & $n_\text{online}$ & \num{100} & \num{100} & \num{100} & \num{100} & Number of tree search iterations of MCTS.\\
              & $c$ & \num{50} & \num{50} & \num{50} & \num{50} & PUCT exploration constant. \\
              & $k_a$ & $\cdot$ & \num{5.0} & $\cdot$ & $\cdot$ & Multiplicative action progressive widening value.\\
              & $\alpha_a$ & $\cdot$ & \num{0.9} & $\cdot$ & $\cdot$ & Exponential action progressive widening value.\\
              & $k_b$ & $\cdot$ & \num{1.0} & \num{1.0} & \num{1.0} & Multiplicative belief-state progressive widening value.\\
              & $\alpha_b$ & $\cdot$ & \num{0.0} & \num{0.0} & \num{0.0} & Exponential belief-state progressive widening value.\\
              & $d$ & \num{15} & \num{15} & \num{4} & \num{4} & Maximum tree depth. \\
              & $\tau$ & \num{1.0} & \num{0} & \num{1.5} & \num{0} & Exploration temperature for final root node action selection. \\
              & $z_q$ & \num{1} & \num{0.4} & \num{1} & \num{0.5} & Influence of $Q$-values in final criteria. \\
              & $z_n$ & \num{1} & \num{0.9} & \num{1} & \num{0.8} & Influence of visit counts in final criteria. \\
            \arrayrulecolor{black} 
            \bottomrule
        \end{tabular}
        \end{footnotesize}
    \end{threeparttable}
    \end{adjustbox}
    \caption{\textit{BetaZero} parameters for the $\textsc{RockSample}$ problems.}\label{tab:rs_params}
\end{table*}

\begin{table*}[pb!]
    \centering
    \begin{adjustbox}{max width=\textwidth}
    \begin{threeparttable}
        \begin{footnotesize}
        \begin{tabular}{@{}clrrm{9cm}@{}}
            \arrayrulecolor{black} 
            \toprule
             & Parameter  &  Offline & Online  &  Description \\
            \midrule
            \multirow{3}{*}{\shortstack{BetaZero policy\\iteration parameters}} & $n_\text{iterations}$ & \num{20} & --- & Number of offline BetaZero policy iterations. \\
             & $n_\text{data}$ & \num{100} & --- & Number of parallel MCTS data gen. episodes per policy iteration. \\
             & bootstrap $Q_0$ & false & false & Use bootstrap estimate for initial $Q$-value in MCTS. \\
            \arrayrulecolor{lightgray}
            \midrule
            \multirow{3}{*}{\shortstack{Neural network\\parameters}} & $n_\text{epochs}$ & \num{10} & --- & Number of training epochs. \\
             & $\alpha$ & \num{1e-6} & --- & Learning rate. \\
             & $\lambda$ & \num{1e-4} & --- & $L_2$-regularization parameter. \\
             \midrule
             \multirow{10}{*}{\shortstack{MCTS\\parameters}} & $n_\text{online}$ & \num{50} & \num{50} & Number of tree search iterations of MCTS.\\
              & $c$ & \num{57} & \num{57} & PUCT exploration constant. \\
              & $k_a$ & \num{41.09} & \num{41.09} & Multiplicative action progressive widening value.\\
              & $\alpha_a$ & \num{0.57} & \num{0.57} & Exponential action progressive widening value.\\
              & $k_b$ & \num{37.13} & \num{37.13} & Multiplicative belief-state progressive widening value.\\
              & $\alpha_b$ & \num{0.94} & \num{0.94} & Exponential belief-state progressive widening value.\\
              & $d$ & \num{5} & \num{5} & Maximum tree depth. \\
              & $\tau$ & \num{1.0} & \num{0} & Exploration temperature for final root node action selection. \\
              & $z_q$ & \num{1} & \num{1} & Influence of $Q$-values in final criteria. \\
              & $z_n$ & \num{1} & \num{1} & Influence of visit counts in final criteria. \\
            \arrayrulecolor{black} 
            \bottomrule
        \end{tabular}
        \end{footnotesize}
    \end{threeparttable}
    \end{adjustbox}
    \caption{\textit{BetaZero} parameters for the $\textsc{Mineral Exploration}$ problem.}\label{tab:minex_params}
\end{table*}

The MCTS parameters for the mineral exploration problem were tuned using Latin hypercube sampling based on the lower-confidence bound of the returns.
During training, the rock sample problems disabled progressive widening to expand on all actions and transition to a single belief state.
Then for online execution, we tuned the DPW parameters as shown in \cref{fig:spw_sensitivity_ld10,fig:apw_sensitivity_ld10,fig:spw_sensitivity_rs20,fig:apw_sensitivity_rs20}.
The problems train with a batch size of $1024$ over $80\%$ of $100{,}000$ samples from one round of data collection (${n_\text{buffer}=1)}$ using $p_\text{dropout}$ of $0.2$, $0.5$, $0.7$, respectively.
The neural network optimizer Adam \citep{kingma2014adam} was used in \textsc{LightDark}$(y)$ while RMSProp \citep{hinton2014rmsprop} was used in the others.
A value function loss of MAE was used in mineral exploration (MSE otherwise), each using $n_\text{samples}=100{,}000$ during training.

\section{Compute Resources}

BetaZero was designed to use a single GPU to train the network and parallelize MCTS evaluations across available CPUs.
Evaluating the networks on the CPU is computationally inexpensive due to the size of the networks (see \cref{fig:nn_lightdark,fig:nn_rocksample,fig:nn_minex}).
This design was chosen to enable future research without a computational bottleneck.
For network training, a single NVIDIA A100 was used with 80GB of memory on an Ubuntu 22.04 machine with 500 GB of RAM.
Parallel data collection processes were run on $50$ processes split evenly over two separate Ubuntu 22.04 machines: (1) with 40 Intel Xeon 2.3 GHz CPUs, and (2) with 56 Intel Xeon 2.6 GHz CPUs.
\Cref{alg:collect_data} (line \ref{line:parallel}) shows where CPU parallelization occurs.
In practice, the MCTS data generation simulations are the bottleneck of the offline component of BetaZero and not the network training---thus, parallelization is useful.

\section{Open-Sourced Code and Experiments}

The BetaZero algorithm has been open sourced and incorporated into the Julia programming language POMDPs.jl ecosystem \citep{pomdps_jl}.
Fitting into this ecosystem allows BetaZero to access existing POMDP models and can easily be compared to various POMDP solvers.
The user constructs a {\color{juliafunccolor}\texttt{BetaZeroSolver}} that takes parameters for policy iteration and data generation, parameters for neural network architecture and training, and parameters for MCTS (described in the tables above).
The user may choose to define a method that inputs the belief $b$ and outputs the belief representation $\tilde{b}$ used by the neural network (the default computes the belief mean and std).
Given a \texttt{pomdp::{\color{juliafunccolor}POMDP}} structure, a \texttt{solver::{\color{juliafunccolor}BetaZeroSolver}} is constructed and solved using:
\begin{equation*}
\texttt{policy = {\color{juliafunccolor}solve}(solver, pomdp)}
\end{equation*}
which runs \textit{offline} policy iteration (\cref{alg:betazero}). Once you have a trained neural network, an action can then be generated \textit{online} from the policy given a belief $b$ using the following (\cref{alg:mcts-top-lvl}):
\begin{equation*}
\texttt{a = {\color{juliafunccolor}action}(policy, b)}
\end{equation*}
All experiments, including the experiment setup for the baseline algorithms with their heuristics, are included for reproducibility.
Code to run MCTS data collection across parallel processes is also included.
The code and experiments presented in this work are available online.\footnote{\url{https://github.com/sisl/BetaZero.jl}}

\section{BetaZero Algorithm}

The following \cref{alg:betazero,alg:collect_data,alg:mcts-top-lvl} detail the full BetaZero policy iteration algorithm that iterates between \textit{policy evaluation} and \textit{policy improvement} for a total of $n_\text{iterations}$.
The offline policy evaluation stage, or data collection process (\cref{alg:collect_data}), runs $n_\text{data}$ parallel MCTS simulations over the original POMDP and collects a dataset $\mathcal{D}$ of beliefs $b_t$, policy vectors $\bpi_t$, and returns $g_t$ (computed after each episode terminates).
The top-level $Q$-weighted MCTS algorithm is shown in \cref{alg:mcts-top-lvl}, which iteratively runs MCTS simulations for $n_\text{online}$ iterations to a specified depth $d$.
The final root node action selection policy follows the $Q$-weighted visit counts from \cref{eq:policy_q_weight}.
The descriptions of parameters $\psi$ used in offline training and online tree search are listed in \cref{tab:ld_params,tab:rs_params,tab:minex_params}.

\begin{figure}[!p]
    \centering
    \input{algorithms/betazero}
\end{figure}

\begin{figure*}[!p]
    \centering
    \begin{minipage}{\textwidth}
        \begin{algorithm}[H]
    \small
    \caption{Collect MCTS data offline for policy evaluation.} 
    \label{alg:collect_data}
    \begin{algorithmic}[1]
        \Function{CollectData$(\mathcal{P}, f_\theta, \psi)$}{}
        \State $\mathcal{D} = \emptyset$
        \ParallelFor{$i \leftarrow 1 \textbf{ to } n_\text{data}$} \GrayComment{parallelize MCTS runs across available CPUs} \label{line:parallel}
            \For {$t \leftarrow 1 \textbf{ to } T$}
                \State $a_t \leftarrow \textproc{MonteCarloTreeSearch}(\mathcal{P}, f_\theta, b_t, \psi)$ \GrayComment{select next action through online planning}
                \State $\mathcal{D}_i^{(t)} \leftarrow \mathcal{D}_i^{(t)} \cup \big\{(b_t, \bpi_\text{tree}^{(t)}, g_t) \big\}$ \GrayComment{collect belief and policy data (placeholder for returns)}
                \State $s_{t+1} \sim T(\cdot \mid s_t, a_t)\phantom{\big\}}$
                \State $o_t \sim O(\cdot \mid a_t, s_{t+1})$
                \State $b_{t+1} \leftarrow \textproc{Update}(b_t, a_t, o_t)$
                \hspace*{3em}%
                \rlap{\raisebox{\dimexpr.5\normalbaselineskip+.5\jot}{\smash{$\left.\begin{array}{@{}c@{}}\\{}\\{}\\{}\end{array}\color{commentgray}\right\}%
                \color{commentgray}\begin{tabular}{l}transition the original POMDP\end{tabular}$}}}
                \State $r_t \leftarrow R(s_t, a_t)$ or $R(s_t, a_t, s_{t+1})$
            \EndFor
            \State $g_t \leftarrow \sum_{k=t}^T \gamma^{(k-t)} r_k$ \textbf{ for } $t \leftarrow 1 \textbf{ to } T$ \GrayComment{compute returns from observed rewards}
        \EndParallelFor
        \State \Return $\mathcal{D}$
    \EndFunction
    \end{algorithmic}
\end{algorithm}
    \end{minipage}
\end{figure*}

\begin{figure*}[!p]
    \centering
    \begin{minipage}{\textwidth}
        \input{algorithms/mcts-outer}
    \end{minipage}
\end{figure*}

\end{document}